\documentclass[pdflatex,sn-apa,iicol]{sn-jnl}

\usepackage{graphicx}%
\usepackage{multirow}%
\usepackage{amsmath,amssymb,amsfonts}%
\usepackage{amsthm}%
\usepackage{mathrsfs}%
\usepackage[title]{appendix}%
\usepackage{xcolor}%
\usepackage{textcomp}%
\usepackage{manyfoot}%
\usepackage{booktabs}%
\usepackage{algorithm}%
\usepackage{algorithmicx}%
\usepackage{algpseudocode}%
\usepackage{listings}%

\theoremstyle{thmstyleone}%
\theoremstyle{thmstyletwo}%

\theoremstyle{thmstylethree}%

\begin{document}

\title[Article Title]{MIGT: Memory Instance Gated Transformer Framework for Financial Portfolio Management}


\author[1,2]{\fnm{Fengchen} \sur{Gu}}\email{Fengchen.Gu19@student.xjtlu.edu.cn}

\author[1]{\fnm{Angelos} \sur{Stefanidis}}\email{Angelos.Stefanidis@xjtlu.edu.cn}
\author[2]{\fnm{Ángel} \sur{García-Fernández}}\email{angel.garcia-fernandez@liverpool.ac.uk}

\author*[1]{\fnm{Jionglong} \sur{Su}}\email{Jionglong.Su@xjtlu.edu.cn}

\author*[1]{\fnm{Huakang} \sur{Li}}\email{Huakang.Li@xjtlu.edu.cn}

\affil*[1]{\orgdiv{School of AI and Advanced Computing, XJTLU Entrepreneur College (Taicang)}, \orgname{Xi’an Jiaotong-Liverpool University}, \orgaddress{\street{Taicang Road}, \city{Taicang, Suzhou}, \postcode{215400}, \state{Jiangsu}, \country{China}}}

\affil[2]{\orgdiv{Department of Electrical Engineering and Electronics}, \orgname{University of Liverpool}, \orgaddress{\street{Brownlow Hill}, \city{Liverpool}, \postcode{L69 7ZX}, \state{Merseyside}, \country{United Kingdom}}}


\abstract{Deep reinforcement learning (DRL) has been applied in financial portfolio management to improve returns in changing market conditions. However, unlike most fields where DRL is widely used, the stock market is more volatile and dynamic as it is affected by several factors such as global events and investor sentiment. Therefore, it remains a challenge to construct a DRL-based portfolio management framework with strong return capability, stable training, and generalization ability. This study introduces a new framework utilizing the Memory Instance Gated Transformer (MIGT) for effective portfolio management. By incorporating a novel Gated Instance Attention module, which combines a transformer variant, instance normalization, and a Lite Gate Unit, our approach aims to maximize investment returns while ensuring the learning process's stability and reducing outlier impacts. Tested on the Dow Jones Industrial Average 30, our framework's performance is evaluated against fifteen other strategies using key financial metrics like the cumulative return and risk-return ratios (Sharpe, Sortino, and Omega ratios). The results highlight MIGT's advantage, showcasing at least a 9.75$\%$ improvement in cumulative returns and a minimum 2.36$\%$ increase in risk-return ratios over competing strategies, marking a significant advancement in DRL for portfolio management.}

\keywords{Portfolio Management, Deep reinforcement learning, Transformer, Stock Market.}



\maketitle

\section{Introduction}\label{s1}

Portfolio management involves determining the allocation of funds to multiple financial assets and continuously changing the portfolio weights to maximize the investment returns \citep{Avramov2002}. The financial investment market is volatile because it is influenced by economic, political, and technological factors \citep{Poon2003}. The financial market data are complex, often consisting of a large volume of historical information such as price, volume, and technical indicators \citep{Liu2021}. The varying conditions of the market environment together with its inherent data complexity, makes it difficult for traditional investment methods to effectively analyze and inform investor decisions \citep{Gao2020, Ren2021}.

Deep learning is a branch of machine learning that uses multiple layers of neural networks to learn representations of data with higher levels of abstraction  \citep{2015Deep}. It has revolutionized the fields of computer vision, speech recognition, and natural language processing \citep{2013Deep, Xia2020Fully}. Reinforcement learning is a machine learning technique in which an agent learns how to operate in an environment by trial and error while maximizing rewards and minimizing penalties \citep{1998Reinforcement, Charpentier2020Reinforcement}. It allows machines and software agents to discover optimal strategies within a specific context automatically \citep{2016Neural}.

Deep reinforcement learning (DRL), combining the perception capabilities of deep learning with decision-making using current knowledge of reinforcement learning, is a practical approach to automated portfolio management \citep{Jiang2017}. The DRL framework learns using feedback from a complex environment and is well suited to addressing dynamic decision-making problems. In portfolio management, stock trading is about making dynamic trading decisions in a complex stock market environment, i.e., deciding when and what to trade, at what price, and the transaction volume \citep{Jiang2018, Yao2022}. 

Recently, researchers applied DRL to portfolio management in the cryptocurrency markets and the stock markets. The cryptocurrency market is particularly volatile in the short term, and high-frequency trading can lead to short-term gains. \citet{Jiang2018} used the Ensemble of Identical Independent Evaluators (EIIE) and policy gradients for cryptocurrency portfolio management. Others subsequently constructed the EIIE framework by modifying the neural networks and adding features as input \citep{Gu2021, Sun2021, Yang2022, Qin2022}. The stock market tends to trade on a single day or over a longer period of time, requiring a long-term portfolio management strategy \citep{Gao2020}. The relatively early framework Investor-Imitator (IMIT) uses reinforcement learning to mimic investor behavior to extract trading knowledge rather than directly using reinforcement learning for portfolio management \citep{2018Investor}. For compatibility with more reinforcement learning algorithms, the FinRL framework presented by \citet{Liu2021} provides reinforcement learning-based stock trading strategies that support single stock trading and portfolio management containing multiple stocks. It supports a variety of reinforcement learning algorithms, e.g., Deep Deterministic Policy Gradient (DDPG) \citep{2018CONTINUOUS}, Soft Actor-Critic (SAC) \citep{2018Soft}, and Proximal Policy Optimization (PPO) \citep{Schulman2017}. Based on FinRL framework, the Ensemble Strategy (ES) uses the DDPG, SAC, and PPO that are selectively applied to different time intervals according to some decision rule \citep{Yang2020}. In addition to FinRL, there is also TradeMaster, an open-source platform for quantitative trading through reinforcement learning, which covers the entire process of designing, implementing, evaluating, and deploying reinforcement learning-based algorithms \citep{sun2023trademaster}. Other works such as SARL augment asset information with price movement prediction and are based on reinforcement learning \citep{Ye2020ReinforcementLearningBP}.

While existing frameworks demonstrate the efficacy of reinforcement learning in portfolio management, some limitations remain. First, the stock market is a complex and volatile environment, with significant fluctuations occurring during trading. This makes training challenging in converging to obtain a reliable policy and update asset portfolio weights effectively \citep{Jos2019, Guan2021}. This may lead to the inability of DRL to train effective strategies to rationally allocate portfolios, resulting in large fluctuations in returns. Second, the existing DRL portfolio management frameworks have insufficient generalization capabilities \citep{Cui2022, Koziarski2020}. Portfolio management should focus more on long-term trends and withstand the impact of short-term volatility, such as "Black swan" events that lead to sharp fluctuations in stock prices. Outliers from such events and other causes can disrupt reinforcement learning's modeling of the normal pattern of stock price movements, leading to the learning of inaccurate associations. Reinforcement learning models may fail to deal with the different features of different market situations, reducing the generalization ability of the strategy \citep{Onireti2016, Kandanaarachchi2020}. Third, the profitability of existing portfolio management models targeting stock markets is not optimal, primarily due to the lack of processing measures for the characteristics of stocks \citep{Gao2022, Soleymani2020}. They contain simple networks with fewer layers or existing networks from other domains without sufficient optimization for portfolio management. In either case, they are unable to adequately handle complex stock data, leading to limited feature extraction capabilities for capturing complex and non-linear relationships in stock data \citep{Zhang2022, Liu101145}. 

The key contributions of our work are threefold. First, the stability of DRL training is enhanced by using the newly constructed Lite Gate Unit (LGU) gating layer as the fan-in layer \citep{Dai2020}, which allows the training to converge smoother and faster, resulting in more effective portfolio management strategies. Second, we propose the use Instance Normalization to balance the scale difference between different feature dimensions of each state, avoiding certain feature dimensions that dominate the model training. Such an approach can reduce the negative impact of single sample outliers and short-term volatility, allowing the network to learn more discriminative feature representations, and improve the generalization ability of the model \citep{Ulyanov2014, Kandanaarachchi2020}. Third, we construct a new Transformer variant (Gated Instance Attention module) to handle complex stock data by leveraging the ability of the model to capture long-range dependencies as well as dependencies among different periods in stock data, which improves the profitability of the DRL portfolio management strategies \citep{Vaswani2017}. It achieves parallel computation through a multi-headed self-attention mechanism, which enables it to efficiently process high-dimensional features of stock data, and eventually improve the profitability of the portfolio management strategies.

The remainder of this paper is organized as follows. In Section \ref {s2}, we define the trading period, make some assumptions about the experimental setting, and give the portfolio management objectives. Section \ref {s3} presents the DRL environment for portfolio management. In Section \ref {s4}, the Memory Instance Gated Transformer (MIGT) policy network is presented. In Section \ref {s5}, we conduct the comparative and ablation experiments as well as offer an interpretation of the results. Finally, conclusions and future work are given in Section \ref {s6}.

\section{Definition}\label{s2}
\subsection{Definition of Data Input, Trading Period and Process}\label{}

\begin{figure*}
	\centering
	\includegraphics[width=14.5cm]{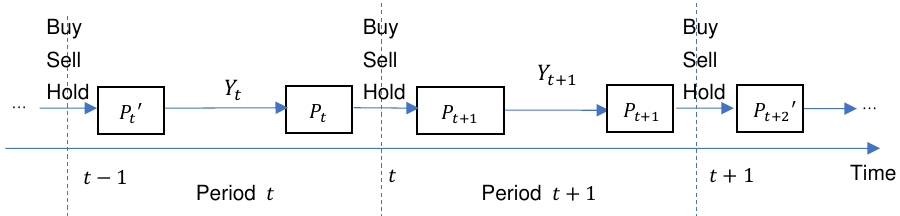}
	\caption{Interaction Process for Time Series}\label{fig1}
\end{figure*}

Under the DRL framework, the agent reallocates capital into different asset classes at each period $t$, $t\in \mathbb{N}^{+}$. We first consider the process of portfolio management in Figure \ref {fig1}. The input to the portfolio management model is a high-dimensional tensor with dimensions of historical time, stock assets, and historical data including price and technical indicators. We set the length of each trading period $t$ to one day. At the end of each period $t$, the agent trades on the portfolio with ${\ V}_t$, the vector of closing prices of all assets in period $t$, and the portfolio value is $P_t$. Portfolio value is the total value of all the securities in the portfolio and the cash sum, i.e., to obtain the total value of the entire portfolio. 

\subsection{Assumptions about the Experimental Environment}\label{}
In order for the constructed trading environment to be as realistic as possible, the following assumptions are made:

\begin{itemize}
	\item Because the trading simulation is based on historical data, we assume our transactions will not affect the price. This is due to the amount traded being tiny relative to the market's overall size, and as such, the impact on stock prices is negligible \citep{Soleymani2020, Jiang2018}.
	\item Since the trading frequency is set to one day, the price of each trade is the previous day's adjusted closing price \citep{Liu2021}. Based on the actions in period $t$, we define the trading actions as sell, buy, and hold. Trading actions by agents are limited to the balance in the account and do not include buying and selling short. 
	\item Different transaction costs are incurred in the stock market, such as trading and execution fees. For compatibility with different trading situations, we assume that our transaction fee rate $c$ is 0.1$\%$ of the value of each trade (buy or sell) \citep{Liu2021}.
\end{itemize}  

\subsection{Portfolio Management Objective}\label{}

The objective of our strategy is to maximize the return of the final portfolio. We define $P_t\in \mathbb{R}^+$ to be the portfolio value at the end of period $t$. The variable $P_t$ contains the available cash $A_t$ and the stock values $E_t$. The available cash in the portfolio at this period is:
\begin{equation}
	A_t=\,\,A_{t-1}+\,\,M_t^TV_t^{\,\,}\,\,\left( 1-c \right) -\,\,B_t^TV_t^{\,\,}\left( 1+c \right), \label{e2}
\end{equation}
where $M_t$ is the share of stock sold in period $t$, $B_t$ represents the proportion of stock bought in period $t$, and $V_t$ is the closing price vector in period $t$. The dimensions of the vectors $V_t$, $M_t$ and $B_t$ are the number of stocks $n$ in the portfolio. The portfolio share vector $W_t$ represents the share of each asset in the portfolio in period $t$. At each time period t, the portfolio share vector $W_t$ is $W_t-1$ from the previous time period adds the shares purchased $B_t$ and subtracts the shares sold $M_t$,
\begin{equation}
	W_t= W_{t-1}+\,\,B_t-\,\,M_t. \label{e3-}
\end{equation}
The value of stock assets $E_t$ at period $t$ is the product of the transposition of the portfolio share vector $W_t$, where the share of each stock is non-negative, and the vector of all assets‘ closing prices $V_t$:
\begin{equation}
	E_t=\,\,W_t^TV_t^{\,\,}=\,\,\left( W_{t-1}+\,\,B_t-\,\,M_t \right) ^TV_t^{\text{
	}}. \label{e3}
\end{equation}
The portfolio's value in period $t$, $P_t$, can be obtained by summing the equations (\ref {e2}) and (\ref {e3}), i.e.,
\begin{equation}
\begin{split}
	P_t =\ A_t+\ E_t \\
        &=\ A_{t-1}+\ {W_{t-1}}^TV_t-\ c{\ \left(B_t+M_t\right)}^TV_t.\label{e4}
\end{split}
\end{equation}
From equation (\ref {e2}), we obtain the value of the stocks in the portfolio in the last period $t-1$,
\begin{equation}
\begin{split}
	P_{t-1}&=\ A_{t-1}+\ E_{t-1}\\&=\ A_{t-1}+\ {W_{t-1}}^TV_{t-1}. \label{e5}
\end{split}
\end{equation}
Subtracting equation (\ref {e5}) from equation (\ref {e4}), we obtain the change in the portfolio value in period $t$, i.e.,
\begin{equation}
\begin{split}
	\bigtriangleup P_t&=P_t-P_{t-1}\\&=W_{t-1}^T\left( V_t-V_{t-1} \right) -c\left( B_t+M_t \right) ^TV_t. \label{e6}
\end{split}
\end{equation}
The change in portfolio value in period $t$ is the increase or decrease in returns at that period, and the goal of portfolio management is to maximize positive returns $\bigtriangleup P_t$ or minimize negative losses during each period $t$.

\section{DRL Environment}\label{s3}
\subsection{The Markov decision process of portfolio management}\label{}

\begin{figure}
	\centering
	\includegraphics[width=8cm]{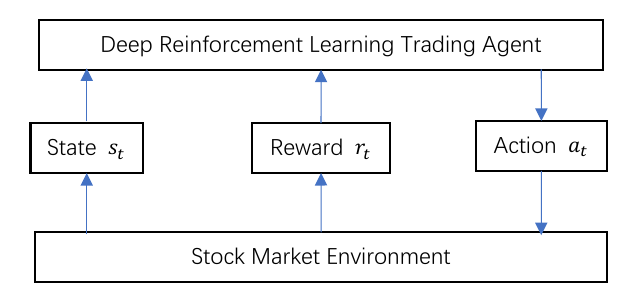}
	\caption{The Markov decision process of the DRL Environment, reflects the interaction of state, reward, action, DRL agents and the stock market environment.}\label{fig2}
\end{figure}

The stock market is stochastic, i.e., the movement of stock prices is random and unpredictable \citep{M.2022Recurrence}. We model the portfolio management task as a Markov decision process problem (Figure \ref {fig2}) to choose an action based on the current state and then randomly move to a new state \citep{Liu2021, Baxter1995}. The training process involves observing changes in state $s_t$, taking actions $a_t$, and calculating rewards $r_t$, allowing the agent to adjust its strategy during the training process. Through this iterative reward-driven learning and feedback process, the DRL agent can train the optimal trading strategy that generates the highest returns for a given environment.
\subsection{State, Action, Reward Function, and transaction agent}\label{}

The state $s_t$ is a representation of a particular situation or context that describes the environment \citep{Naeem2020A}. Transaction agents observe many features in an interactive market environment to make sequential decisions \citep{Arulkumaran2017}. To make the model adaptive to stock trends and volatility data, we use different features to help agents to make informed decisions and use them to construct the state. Previous frameworks have generally used the following items as states:
\begin{itemize}
	\item $A_t$, available cash assets of stocks in period $t$;
	\item $W_t$, the share vector of each stock in the portfolio in period $t$; and
	\item $V_t$, the closing price vector in period $t$.
\end{itemize} 

To help portfolio management strategies identify uptrends and downtrends in each stock, the following trend indicators are  included as part of the state: 
\begin{itemize}
	\item Bollinger bands (BOLL) can be used to determine the range of stock price fluctuations and future movements, and to show the safe high and low-price levels of a stock \citep{Murphy1999};
	\item Commodity Channel Index (CCI) mainly measures the variability out of the normal range of prices \citep{Altan2022};
	\item Relative Strength Index (RSI) is a momentum indicator that measures the speed and magnitude of price movements \citep{Altan2022}; and
	\item True Range of Trading (TR) is used to measure the intensity of market volatility \citep{Chang2019}.
\end{itemize} 
Overbought and oversold indicators can reflect the short-term volatility of each stock and help portfolio management strategies perceive risk. Therefore we also include the following volatility indicators: 
\begin{itemize}
	\item Directional Movement Index (DMI) is used to determine the movement of stock prices by analyzing the change in the equilibrium point between buyers and sellers during the rise and fall of stock prices \citep{Seyma2020};
	\item Moving Average Convergence Divergence (MACD)  is a technical indicator that uses the convergence and divergence between the short-term exponential moving average and the long-term exponential moving average of the closing price to make a judgment on the timing of a trade  \citep{Hung2016}; and
	\item Money Flow Index (MFI)  is a technical indicator that uses trading volume and price to determine overbought and oversold \citep{Singleton2014}.
\end{itemize} 
The state tensor in period $t$ is $s_t=\left[f_{1,t},\ f_{2,t},\ \ldots,\ f_{n,t}\right]$, where $n$ the number of stocks in the portfolio. Each element of $s_t$, $f_{k,t}=\left[A_{k,t},\ W_{k,t},\ V_{k,t},\  BOLL_{k,t},\ \ldots,\ MFI_{k,t}\right]$, is the feature vector of $k$-th stock in period $t$, $k=1,2,3,\ldots,\ n$.

To allow the model to deal directly with portfolios rather than individual stock trades, the process of transaction is the portfolio in period $t$ for each asset in $E_t$ after a buy, sell or hold operation \citep{Poon2003, Kumar2021A}. The action $a_t$ is a portfolio share vector $W_t$ that reflects the buying and selling behavior of the transaction agent.

The role of the reward function $r\left(s_t,a_t,\ s_{t+1}\right)$ is to define the goals of DRL and evaluate the transaction agent’s behavior \citep{Lehnert2020Reward-predictive, Arulkumaran2017}. The change in the portfolio value in period $t$ from the previous period $t-1$ is presented in equation (\ref {e6}). Therefore, the reward in period $t$ is:
\begin{equation}
\begin{split}
	r\left( s_t,a_t,s_{t+1} \right) =\bigtriangleup P_t &=W_{t-1}^T\left( V_t-V_{t-1} \right) \\&-c\left( B_t+M_t \right) ^TV_t.
\end{split}
\end{equation}
PPO \citep{Schulman2017} controls the policy gradient update and ensures that the new policy will be numerically close to the previous one so that it can learn incrementally without destabilizing its own learning process. It uses an objective function:

\begin{equation}
\begin{split}
L^{CLIP}\left( \theta \right) = \hat{\mathbb{E}}_t [ &\min \left( r_t\left( \theta \right) \right) \hat{A}_t,\\&clip\left( r_t\left( \theta \right) ,1-\epsilon ,1+\epsilon \right) \hat{A}_t ], 
\end{split}
\end{equation}
where $\theta$ is the policy parameter, ${\hat{\mathbb{E}}}_t$ denotes the empirical expectation over period $t$, $r_t\left(\theta\right)=\ \frac{\pi_\theta\left({a_t\ |\ s}_t\right)}{\pi_{\theta_{old}}\left({a_t\ |\ s}_t\right)}$ is the probability ratio of the new policy to the old one, ${\hat{A}}_t$ is the estimated advantage function in period $t$, and $clip\left(r_t\left(\theta\right),\ 1\ -\ \epsilon,\ 1\ +\ \epsilon\right) $ truncates the ratio to $r_t\left(\theta\right)$ in the range $[1\ -\ \epsilon,\ 1\ +\ \epsilon]$. The $clip$ operation is used to limit the step size of the policy update to prevent the policy update from exceeding a certain threshold, enabling the PPO to be more stable and reliable \citep{Zhu2021A}.

\section{Policy Network}\label{s4}
\subsection{Research Problem}\label{}
DRL is successfully applied to many domains, especially those with restricted state and action spaces that are conducive to exploration. The environments of most DRL tasks are relatively stable, with rules that persist and do not change frequently, i.e., chess and other confrontational games \citep{Chasparis2012Distributed, Khader2021Learning}. The stock market is challenging for DRL due to its instability and dynamics \citep{Parisi2020Reinforcement}. Portfolio management tasks face a more dynamic and volatile environment than typical DRL application domains, being influenced by many external factors such as the interplay between corporate earnings, macroeconomics, and policies \citep{Nasir2014Aspects, Sharif2020COVID-19, Byrne2021The}. Irrational investor behaviors and information asymmetry further destabilize the market, causing drastic fluctuations in individual stock prices \citep{Paule-Vianez2020A}.

In such a complex and dynamic environment, DRL faces three main challenges:
\begin{itemize}
	\item The stock market environment is always changing, but DRL relies on a stable environment to accumulate experience \citep{Khader2021Learning}. Existing DRL models can hardly deal with high-dimensional, non-linear, and dynamic stock data. They usually use fixed-length information, which can lead to truncation or zeroing of the sequence, destroying the original time-series structure and potentially losing valuable historical data \citep{Cao2019Deep}. In addition, basic DRL models have difficulty effectively learning and modeling the long-term dependencies present in stock time series data, which makes it challenging to understand underlying market trends, and the profitability of portfolio strategies is compromised \citep{Gershman2010Context, Hu2021A}. 
    \item In the context of the stock market, company information, policy environment, investment psychology, etc. in the stock market are changing constantly, making the training of DRL models challenging. The frequent emergence of new information in the stock market leads to the continuous need for adaptation of prices and trading strategies. However, typical DRL is inefficient in adapting to new environments, making it difficult to train a stable and reliable strategy \citep{Dayan2008Reinforcement, Gupta2021Deep}. This is manifested in the difficulty of convergence of the training process and the eventual under-training that leads to poor strategy results \citep{Zhang2021A}. 
    \item There are various types of short-term market disturbances and noisy trades in the stock market, and these disturbances flood the data, interfering with the DRL algorithm's identification of effective signals. Normal fluctuations and "Black swan" incidents can expose DRL susceptible to overfitting problems, resulting in strategies that excel at short-term stage but have poor generalization ability \citep{Song2019Observational, Whiteson2011Protecting, Ale2020Dragons}.
\end{itemize}

\subsection{MIGT Framework}\label{}

\begin{figure}
	\centering
	\includegraphics[width=7cm]{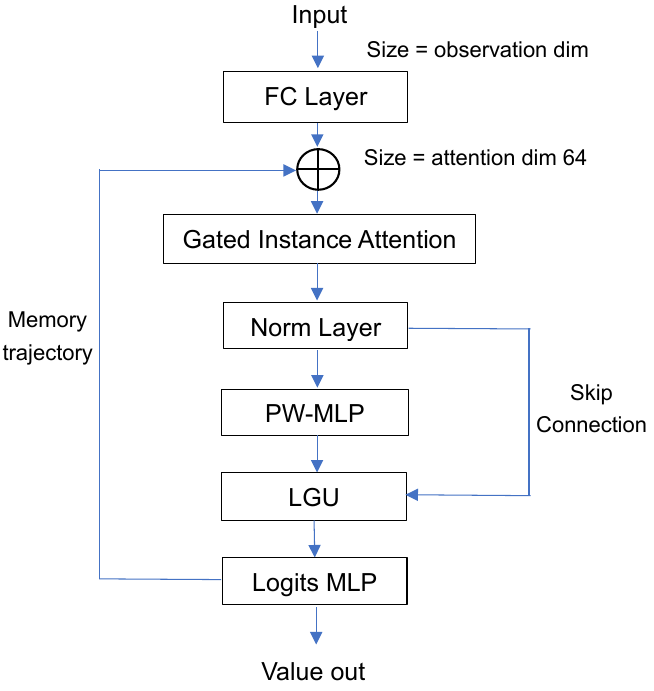}
	\caption{the structure of MIGT policy}\label{fig3}
\end{figure}

To address these challenges, we propose the MIGT Framework (Figure \ref {fig3}), as the neural network for PPO input. The input to the portfolio management model is historical data in a high-dimensional tensor with historical period $t$, stock assets number $n$ and and historical data $f$ comprising price and technical indicators. Each period $t$ is a trading day, so the historical time dimension of the input tensor is the number of periods $t$, which is the number of trading days. The stock assets dimension defines the number of stocks $n$. The input data first passes through a standard FC layer, followed by memory trajectory processing. Subsequently, the tensor is processed by the Gated Instance Attention module (Figure \ref {fig5}). The next module is a Position-wise Multilayer Perceptron (PW-MLP) with normalization and fan-in layer. Finally, the data from the PW-MLP are fed into the Logits Multilayer Perceptron (Logits MLP).

\subsection{Gated Instance Attention}\label{}

\begin{figure}
	\centering
	\includegraphics[width=6cm]{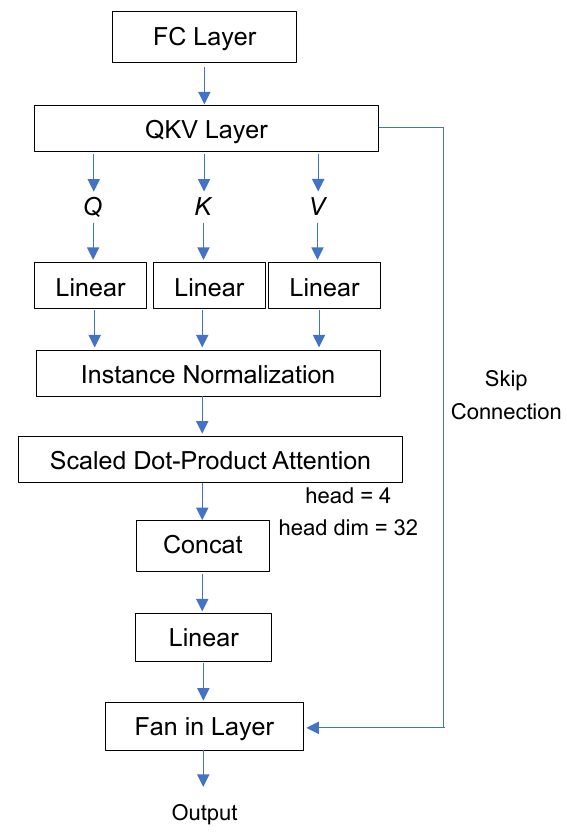}
	\caption{the structure of Gated Instance Attention}\label{fig5}
\end{figure}

\begin{figure}
	\centering
	\includegraphics[width=3.2cm]{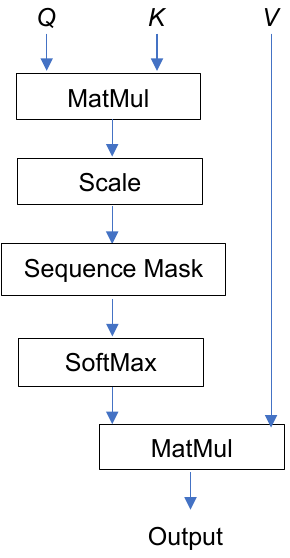}
	\caption{the Scaled Dot-Product Attention structure}\label{fig4}
\end{figure}

Figure \ref {fig5} gives the structure of the Gated Instance Attention. The Transformer's self-attention mechanism can weigh each position in the input sequence to focus on the information of other positions so that the multi-dimensional information in the stock history data can be processed \citep{Vaswani2017}. As given in Figure \ref {fig4}, the input to the scaled dot-product attention module consists of the query vector $Q$ and key vector $K$ of the dimension and the value vector $V$ of the dimension $d_V$ \citep{Vaswani2017}. We compute the dot product of the query and all keys, dividing each key by the root $d_K$, and use the SoftMax function to derive the weights of the values. In this network, we compute the attention function for a set of queries and store them in the query vector $Q$. The keys and values are also stored in a key vector $K$ and a value vector $V$. We compute the output vector as \citep{Vaswani2017}:
\begin{equation}
Attention\left( Q,K,V \right) =\,\,softmax\left( \frac{Q\,\,K^T}{\sqrt{d_K}} \right) \,\,V
\end{equation}
\\where $\sqrt{d_K}$ represents the scaling factor that prevents inflation in calculations from causing unstable values. The correlation between query and key is calculated by using nonlinear matching functions including dot product, which can deal with the nonlinear relationship and the complex dynamic stock market environment. The model receives all available data in a scaled dot-product attention layer, but it only accesses one representation space. We divide the model into multiple heads, forming multiple subspaces so that the model can learn relevant information in different representation subspaces. Through the multi-head mechanism, different features of multi-dimensional stock data can be learned from different subspaces, which helps to deal with high-dimensional inputs. In order to focus on the different parts of the input sequence, we use four heads to perform the attention computation simultaneously, without sharing parameters before each other and eventually stitching the results together \citep{Vaswani2017}. Simultaneous computation of multiple attentions can focus on different parts of the input, which can help to capture the key characteristic information of the stock time series to deal with changes in its dynamics.

Transformers use layer normalization in Natural Language Processing and other fields after calculating self-attention and after the feedforward portion of each encoder block, which is usually applied to the same sample \citep{1214993, Vaswani2017}. In contrast, instance normalization \citep{Ulyanov2014} is used in the attention module as it is applied to each channel data, such as the multi-dimensional time series. Layer Normalization normalizes the whole layer network and loses the specific information of each moment of the time series, while Instance Normalization retains the original differences between each sample, such as the time series information, but only reduces the differences within each sample. Preventing instance-specific mean and covariance shifts mitigates the effect of outliers in stock price and feature data \citep{Kandanaarachchi2020, Ulyanov2014}. Instance normalization reduces the bias of individual samples and allows the network to focus more on global information. This improves the robustness of the model against normal fluctuations and unexpected occurrences, leading to improved overall performance. Instance normalization also facilitates the network to learn more discriminative feature representations, which leads to a further improvement in the return ability of the portfolio management strategy. Therefore, the normalization layer is moved to the front of the attention module so that the normalization layer operation can be applied to the input of the sub-module. The advantage of this reordering is that the reordered network supports the mapping of identities from the input of the first transformer layer to the output after the last layer \citep{Dai2020}. The layer canonical reordering results in a path where the two linear layers are applied sequentially, the non-linear ReLU activation is applied to the output stream.

We propose a novel approach to construct the LGU gating layer ($g$ in function \ref {e12}) as a fan-in layer to replace the remaining connections after the attention module and PW-MLP. Unlike the Gated Recurrent Unit \citep{Chung2014}, we use an update gate $z$, avoiding the offsetting effect of the gated outputs. This simplifies the computation process and reduces the number of derivable parameters, thus resulting in a smoother gradient. We use $sigmoid$ as the activation function in the update gate $z$ and the hidden layer $h$, which has a smooth function derivative curve with non-zero gradients at both ends to avoid vanishing gradients. LGU can be represented by equations (\ref {e10}), (\ref {e11}) and (\ref {e12}), where the update gate $z$ replaces the information in the previous state selectively instead of discarding it,
\begin{equation}
	z=sigmoid\left(W_zy+\ U_zx-\ b_g\right). \label{e10}
\end{equation}
The hidden layer $h$ is used to store the calculation results, which makes it easier for the model to capture information at more distant moments and to learn long-term dependencies:
\begin{equation}
	h=sigmoid\left(W_gy+\ U_g(z\odot x)\right). \label{e11}
\end{equation}
The final calculation result is output by the output layer $g$, i.e.,
\begin{equation}
	g\left( x,y \right) =\left( 1-z \right) \odot x^{\,\,}+z\odot h^{\,\,} , \label{e12}
\end{equation}
where $g$ is a gating layer function, $y$ is the input, $x$ is the residual value, and $\odot$ is the elemental multiplication. $W$ and $U$ are parameters for each gating neuron. Using LGU as gating layer can reduce the possibility of the model learning irrelevant information and, as such, improve the model’s attention to important information, thus improving the stability of learning.

\subsection{PW-MLP, Logits MLP and Memory Trajectory}\label{}

\begin{figure}
	\centering
	\includegraphics[width=6cm]{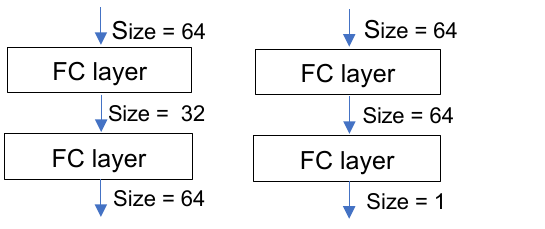}
	\caption{the PW-MLP (left) and Logits MLP (right)}\label{fig6}
\end{figure}

The PW-MLP retains the dimensionality of the inputs and outputs while increasing the expressiveness of the model. As given in Figure \ref {fig6}, it applies two standard FC layers to each input position individually \citep{Dai2020}. The input size of the first layer is attention dimension size 64, and the output dimension is PW-MLP dimension 32. The input and output sizes of the second layer are the opposite of that of the first layer, i.e., its input and output sizes are PW-MLP dimension, and attention dimension, respectively. This allows the PW-MLP to learn non-linear transformations at each position independently, which enhances the model's ability to capture complex relationships in the data, i.e., increasing the expressiveness of the model. The final layer used to output values is the Logits MLP in Figure \ref {fig6}, which consists of two FC layers \citep{Goodfellow2014}. The first layer has the same input dimension as the attention module, and the output outputs have the same size as the attention. The output of the final layer will be used by the PPO algorithm to update the model.

The basic attention operation does not explicitly consider the sequence order because it is reciprocally invariant \citep{Vaswani2017}. As such, we incorporate a novel recursive mechanism into the Transformer architecture to address the limitation of using fixed time length information. In each step of the recursion, the model uses the information of the current time step and the previous time steps to generate output, and the output of the previous step is used as the input of the next step, which continues to generate the output until the end of the sequence. To allow the model to fuse current input and historical memory to deepen its learning ability, we cache a predetermined length of old hidden states across multiple segments and refer to them as memory. In the memory trajectory scheme, there is an additional $t$-step memory tensor $M_t(I,D)$, where $I$ is memory inference and $D$ refers to past memory, whose value is the attention dimension. Memory is used as an additional input to the attention layer during each training session. By using memory states as additional input to the neural network, the attention mechanism can consider current and historical information, improving the model’s inference and comprehension.

\section{Experiments}\label{s5}
\subsection{Data Selection}\label{}

To assess the performance of the MIGT framework, we conduct experiments using historical data, using 30 stocks from the Dow Jones Industrial Average (DJIA) \citep{DJIA} as the portfolio. We use three datasets of calendar years 2019, 2020, and 2021, as the test sets for Experiments 1, 2, and 3, respectively. The training set for each experiment is the data from the three years prior to their test set. The years of these datasets represent the pre-COVID-19, early, and mid-term COVID-19 years, respectively. The specific time composition of the training set and test set are given in Table \ref{tbl1}.

\begin{table*}
        \centering
	\caption{Time composition of the training and test set}\label{tbl1}
	\begin{tabular}{l l l l}
		\toprule
		Dataset & Data purpose& Training data set& Back-test data interval  \\
		\midrule
		1 & Back-test 2019 & 2016.01.01– 2018.12.31 & 2019.01.01– 2019.12.31  \\
		2 & Back-test 2020 & 2017.01.01– 2019.12.31 & 2020.01.01– 2020.12.31  \\
		3 & Back-test 2021 & 2018.01.01– 2020.12.31 & 2021.01.01– 2021.12.31  \\
		\bottomrule
	\end{tabular}
\end{table*}

\subsection{Performance Measures}\label{}
The most direct and effective way to evaluate portfolio management frameworks is to compare cumulative rates. We use the cumulative rate of return as the metric, i.e.,
\begin{equation}
	Cumulative\ rate\ of\ return\ =\ \frac{P_T\ -P_0}{P_0}\,
\end{equation}
where $P_0$ is the initial money of the portfolio, $T$ is the total number of periods $t$ and $P_T$ is its final value. The cumulative rate of return provides a direct representation of the absolute level of return of an investment strategy, which gives how much profit the investment strategy can generate. However, the cumulative rate of return does not reflect the level of risk as it only calculates all periodic returns without considering fluctuations over the investment \citep{Bollerslev2020Good}. The Sharpe ratio \citep{Sharpe1994} is the most widely used risk-return indicator,

\begin{equation}
	Sharpe\ Ratio\ =\ \frac{R_p-R_f}{\sigma_p},
\end{equation}
where $p$ is the portfolio, $R_p$ is the return on the asset, $R_f$ is the risk-free return, which value was set to 3$\%$ per year (the highest level from 2019 to 2021) in our experiments, and $\sigma_p$ is the standard deviation of the asset. This ratio measures the amount of one receives for every unit of risk taken. Therefore, the higher the value of the Sharpe ratio, the higher the portfolio's investment return for the same risk.

When evaluating a portfolio, it is important to consider not only the general level of risk but also the potential for downside risk. The potential for downside risk refers to the possibility of experiencing losses when returns are negative. The Sortino ratio \citep{Rollinger2015} is a risk-adjusted indicator to determine the additional return that an investment generates per unit of downside risk, 
\begin{equation}
	Sortino\ Ratio\ =\ \frac{R_p-r}{\sqrt{\frac{1}{T}\sum_{t=0}^{T}\left(R_{p_t}-r\right)^2}},
\end{equation}
where $r$, whose value was set to 3$\%$ per year in our experiments, is the minimum acceptable return we are considering, $T$ is the total number of periods $t$, and $R_{p_t}$ rate of return value for stock $p$ in period $t$. A higher Sortino ratio means that the investor will receive a higher return per unit of downside risk.

To comprehensively evaluate the risk-return characteristics of the portfolio, we also use the Omega ratio \citep{Keating2002}. It does not depend on the overall distribution of portfolio returns, but is calculated directly using the cumulative distribution function, i.e.,
\begin{equation}
	Omega\,\,Ratio\,\,=\,\,\frac{\int_r^{\infty}{\left( 1-F\left( x \right) \right)}dx}{\int_{-\infty}^r{F}\left( x \right) dx},
\end{equation}
where $F$ is the cumulative distribution function of the returns. This ratio indicates the extent to which the gain component exceeds the loss component and measures the odds of winning versus losing. A higher Omega ratio means that the investment provides a higher return relative to the downside risk assumed.

\subsection{Comparative Strategies}\label{}
To assess the performance of our strategy, traditional statistical strategies and DRL strategies are used as comparative strategies. The traditional statistical strategies we use are based on mean reversion and trend following.

The mean reversion strategies \citep{Fil2020Pairs, Mousavi2021A} aim to take advantage of stock price fluctuations, as stock prices constantly fluctuate over a certain period. The strategies based on mean reversion are:

\begin{itemize}
	\item Confidence Weighted Mean Reversion (CWMR) \citep{Li2011} models the portfolio vector as a Gaussian distribution and updates the distribution sequentially by following the mean reversion trading principle.
	\item Online Moving Average Reversion (OLMAR) \citep{Li2012} uses multi-period moving average regression.
	\item Passive Aggressive Mean Reversion (PAMR) \citep{Lii2012} relies on the mean reversion relationship of financial markets and utilizes online passive-aggressive learning techniques in machine learning.
	\item Robust Median Reversion (RMR) \citep{Huang2016} takes advantage of the mean reversion properties of financial markets and uses a technique called "robust L1-median estimation" to solve the outlier problem in mean reversion.
	\item Transaction costs optimization (TCO) \citep{Li2018} is a strategy for non-zero transaction costs that combines the L1 parametrization of the difference between two consecutive allocations with the principle of maximizing expected logarithmic returns.
	\item Weighted Moving Average Mean Reversion (WMAMR) \citep{Gao2013} is calculated by taking a weighted moving average of stock prices to predict the upward or downward trend of stock prices.
\end{itemize} 

Trend-following strategies \citep{Fousekis2021Returns, Takada2022Trend-following}, on the other hand, aim to take advantage of stock price trends. Strategies based on the trend following are:

\begin{itemize}
	\item The algorithm selects the asset with the best performance on the last day (BEST) \citep{Jiang2017},
	\item Nearest neighbor-based strategy (BNN) \citep{Laszlo2006} uses proximity to classify or predict groups of individual data points.
	\item Correlation-driven nonparametric learning approach (CORN) \citep{Lii2011} is a correlation-based nonparametric learning approach that uses correlations to infer relationships between variables.
\end{itemize} 

 EIIE, IMIT, FinRL, ES, TradeMaster, and SARL are used in our experiments as DRL comparative frameworks. EIIE has performed well in the cryptocurrency portfolio management space \citep{Jiang2017}, and we have migrated their framework to the stock portfolio space and optimized it for the stock market. IMIT formalizes trading knowledge by mimicking investor behavior using a set of logical descriptors and introduces a Rank-Invest model that learns to optimize different evaluation metrics to maintain the diversity of logical descriptors \citep{2018Investor}. FinRL is a well-structured and effective basic framework for automated trading using DRL \citep{Liu2021}. ES combines the best features of three actor-critic-based algorithms and is a novel portfolio management framework \citep{Yang2020}.  In addition to enabling DRL-based quantitative trading (including portfolio management), TradeMaster introduces automated machine learning techniques to tune the hyperparameters that train the reinforcement learning algorithms \citep{sun2023trademaster}, which we will use in conjunction with the PPO algorithm, TradeMaster\_PPO (TMP), for comparison. SARL uses asset information and price trend prediction as additional states to incorporate heterogeneous data and enhance robustness to environmental uncertainty, while price trend prediction can be based on financial data \citep{Ye2020ReinforcementLearningBP}.

\subsection{Comparative Results}\label{Results}

The results of the experiments are given in Table \ref {tbl2}, where the best results for each group are given in bold. The experimental results show that MIGT performs best in all three experiments. The MIGT outperforms the comparative strategies by at least 9.75$\%$, suggesting that our strategy has a stronger ability to capture returns. In terms of the Sharpe ratio, our framework is at least 0.2072 higher than the comparative strategy, indicating that our framework can generate higher risk-adjusted returns per unit level of risk. The Sortino ratio of our strategy is at least 0.3858 higher than the comparative strategy, suggesting that our framework can generate higher excess returns per unit of downside risk. MIGT's Omega ratio remains the highest of the three experiments, with an margin of at least 0.0286 in the experiment, although it has a slight advantage over the other metrics. This demonstrates that, for the same task and dataset, our framework has a higher probability of obtaining positive returns and stronger sustained profitability.

\begin{table*}
        \centering
	\caption{Results of the comparative experiments, where the best results for each metric are in bold.}\label{tbl2}
	
	\begin{tabular}{l l l l l}
		\toprule
		Strategies& Cumulative returns & Sharpe ratio & Omega ratio & Sortino ratio \\  
		\toprule
		Dataset 1\\
		\midrule
		\pmb{MIGT}  &\pmb{0.38443}  & \pmb{1.72498}  &  \pmb{1.33987} &  \pmb{2.74069}       \\
        EIIE & 0.17211 & 0.95176 & 1.19001 & 1.65438 \\
        IMIT  & -0.00403 &	-0.06185 &	0.98952 &	-0.08255 \\
        Finrl & 0.09105 & 0.42895 & 1.07678 & 0.82861 \\
        ES & 0.17291 & 1.35619 & 1.27595 & 2.30452 \\
        TMP & 0.21951 &	1.51779 &	1.30771 &	2.14405 \\
        SARL & 0.20178 &	1.36616 &	1.25836 &	1.93005\\
        BEST & -0.10451 & -0.50668 & 0.91478 & -0.54222 \\
        BNN & -0.42753 & -3.09789 & 0.59106 & -3.54361 \\
        CORN & -0.37279 & -2.38758 & 0.64868 & -2.65702 \\
        CWMR & -0.28261 & -1.28733 & 0.79902 & -1.47289 \\
        OLMAR & -0.32590 & -1.48064 & 0.77743 & -1.72265 \\
        PAMR & -0.28734 & -1.30985 & 0.79574 & -1.49827 \\
        RMR & -0.30868 & -1.39968 & 0.78643 & -1.64853 \\
        TCO1 & -0.03966 & -0.32699 & 0.94653 & -0.20877 \\
        WMAMR & -0.05699 & -0.25328 & 0.95863 & -0.17016 \\
		\toprule
		Dataset 2\\  
		\midrule
		\pmb{MIGT}     & \pmb{0.24440}      & \pmb{0.70689}      & \pmb{1.14642}     & \pmb{1.18089}       \\
		EIIE & 0.14689 & 0.47497 & 1.10464 & 0.79507 \\
        IMIT & -0.25268 &	-0.51091 &	0.90696 &	-0.67185 \\
        Finrl & 0.01942 & 0.19664 & 1.04048 & 0.37425 \\
        ES & 0.05654 & 0.22474 & 1.04322 & 0.50492 \\
        TMP & 0.07834 &	0.30417 &	1.06416 &	0.42602 \\
        SARL & 0.08555 &	0.32463 &	1.06734 &	0.46560 \\
        BEST & -0.44279 & -0.64628 & 0.88068 & -0.90213 \\
        BNN & -0.31112 & -0.48298 & 0.90252 & -0.68164 \\
        CORN & -0.04534 & 0.00672 & 1.00147 & 0.11417 \\
        CWMR & -0.60932 & -1.33623 & 0.74598 & -1.62408 \\
        OLMAR & -0.29166 & -0.21499 & 0.95504 & -0.22308 \\
        PAMR & -0.61574 & -1.35462 & 0.74277 & -1.64354 \\
        RMR & -0.25559 & -0.20466 & 0.95724 & -0.20269 \\
        TCO1 & -0.50058 & -0.97505 & 0.79868 & -1.18684 \\
        WMAMR & -0.15039 & 0.04687 & 1.01031 & 0.12636 \\

		\toprule
		Dataset 3\\  
		\midrule
		\pmb{MIGT}      & \pmb{0.28336}            & \pmb{1.30133}      & \pmb{1.24129}     & \pmb{2.18922}       \\
		EIIE & 0.13370 & 0.54816 & 1.10077 & 0.97863 \\
        IMIT & 0.14955  &	0.76728 &	1.13437 &	1.10643 \\
        Finrl & 0.11807 & 0.56680 & 1.09693 & 1.08441 \\
        ES & 0.16076 & 0.89057 & 1.16002 & 1.63785 \\
        TMP & 0.16901  &	1.16026 &	1.21270 &	1.67973 \\
        SARL & 0.14715 &	0.93386 &	1.16735 &	1.33449 \\
        BEST & -0.37901 & -1.58421 & 0.74931 & -1.90757 \\
        BNN & 0.12454 & 0.48104 & 1.08307 & 0.95923 \\
        CORN & -0.26964 & -1.18858 & 0.80911 & -1.36825 \\
        CWMR & -0.41946 & -2.23410 & 0.67878 & -2.70346 \\
        OLMAR & -0.24008 & -1.07469 & 0.83282 & -1.32658 \\
        PAMR & -0.41617 & -2.21117 & 0.68143 & -2.67586 \\
        RMR & -0.29132 & -1.33161 & 0.79731 & -1.64067 \\
        TCO1 & -0.13407 & -0.92436 & 0.85288 & -1.07723 \\
        WMAMR & -0.21895 & -1.02360 & 0.83832 & -1.27236 \\

		\bottomrule
	\end{tabular}
\end{table*}

\begin{figure}
	\centering
	\includegraphics[width=8.5cm]{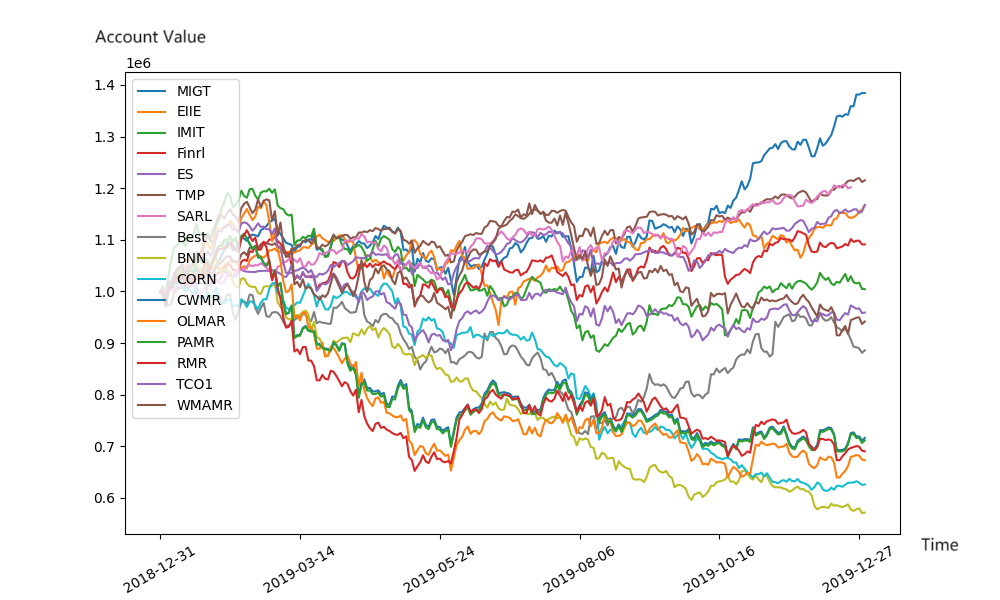}
	\caption{Results of the comparative experiments of data 2019.}\label{figc1}
\end{figure}

\begin{figure}
	\centering
	\includegraphics[width=8.5cm]{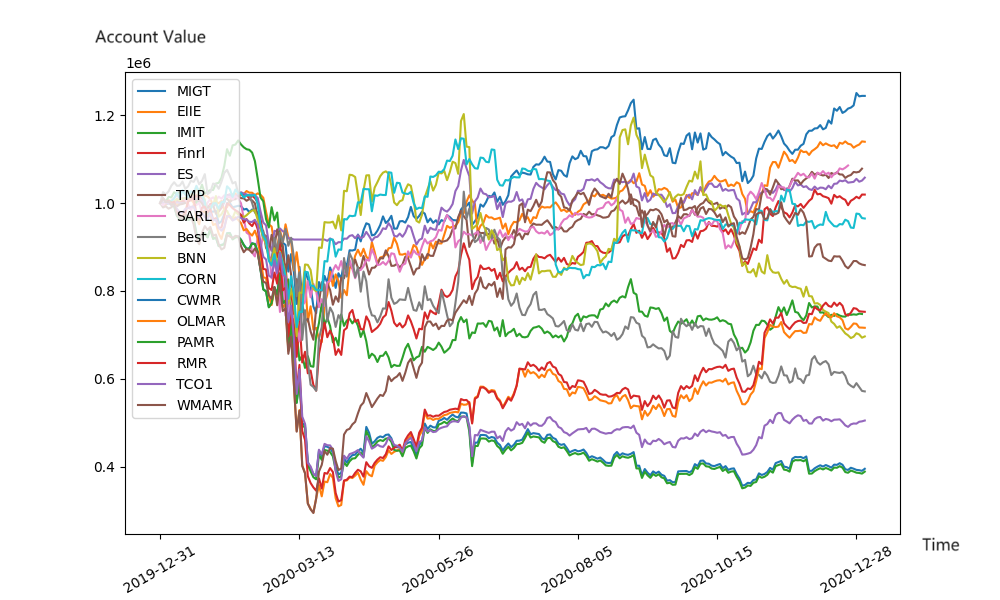}
	\caption{Results of the comparative experiments of data 2020.}\label{figc2}
\end{figure}

\begin{figure}
	\centering
	\includegraphics[width=8.5cm]{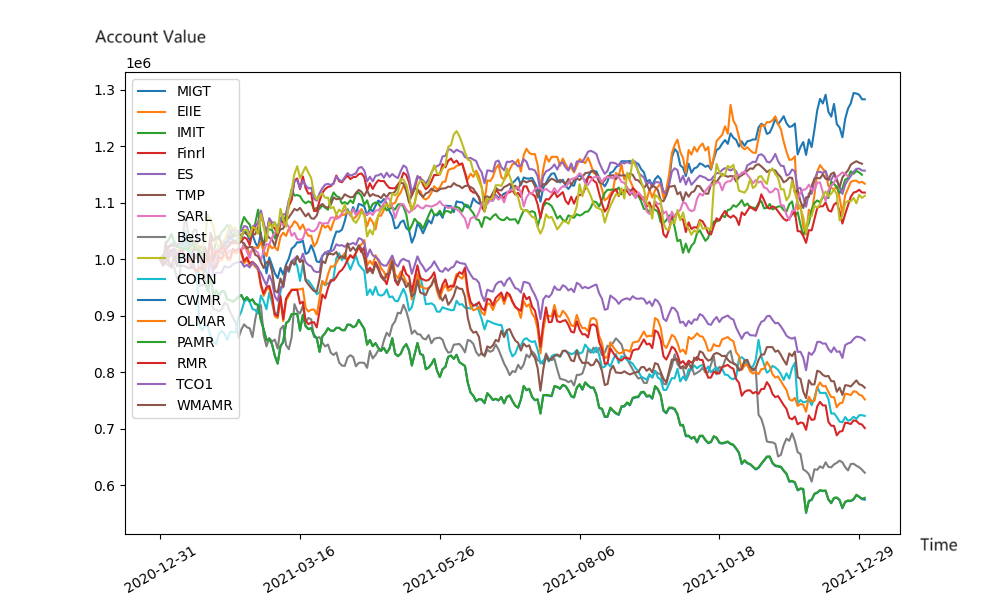}
	\caption{Results of the comparative experiments of data 2021.}\label{figc3}
\end{figure}

Figures \ref {figc1}, \ref {figc2} and \ref {figc3} give the ratios of the portfolio value to the initial value for each day in the experiments. The backtest result of MIGT does not initially exhibit an obvious advantage in the first 2 months, maintaining performance comparable to the top three strategies, as the DRL strategy needs time to adjust, optimize, and obtain the best investment allocation. However, after about 3-8 months of adaptation to the market, the effectiveness starts to become evident. In approximately the last four months, our strategy shows a clear advantage and outperforms the other strategies substantially until the end of the backtest. Our strategy adapts over time to changes and complexities in the market environment through DRL. The initial period where the performance is comparable to other strategies reflects the time required to learn effective policies tailored to the market. Once learned, the DRL strategy enables the portfolio to leverage opportunities and risks, demonstrating robust returns. In contrast, the other strategies appear limited to static rules that fail to fully adapt to the dynamic market, ultimately limiting their performance.

\begin{figure*}
    \centering
    \includegraphics[width=17cm]{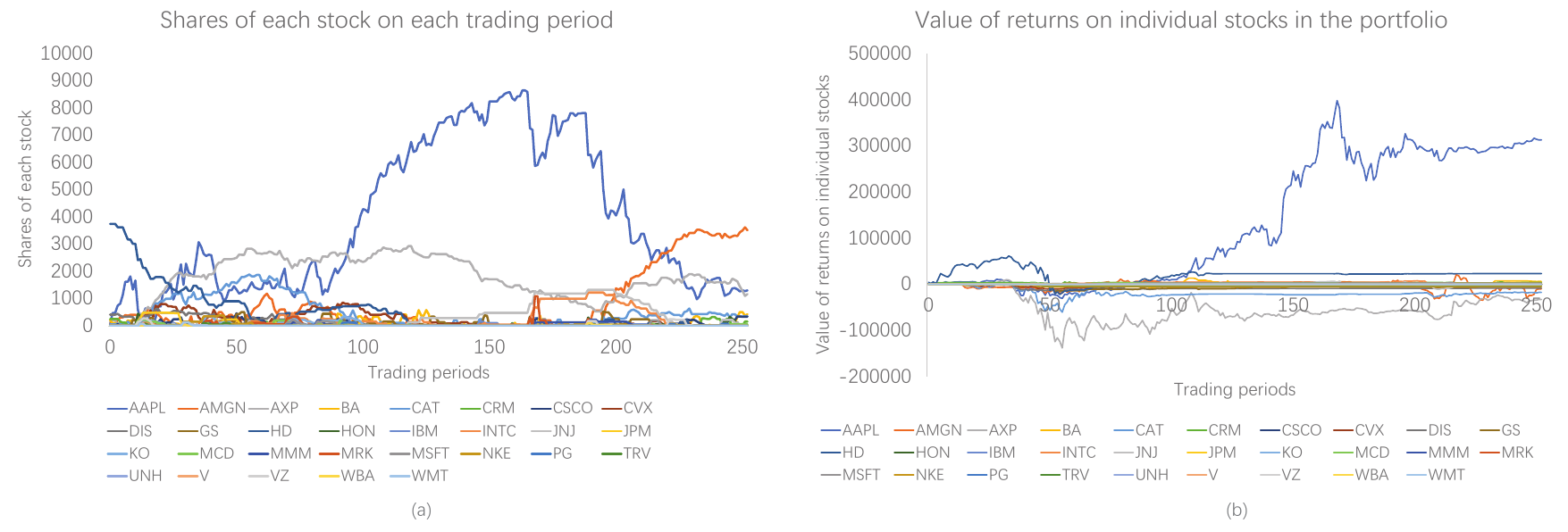}
    \caption{Schematic diagram of the process of adjusting the stock shares by the transaction agent during each trading period $t$, and the gain or loss incurred by each stock in the portfolio.}\label{figr}
\end{figure*}

To further investigate how our MIGT can achieve higher cumulative returns, we visualize the portfolio vector and the daily returns of individual stocks in Figure \ref {figr}, using the dataset 2 as an example. As can be seen in Figure \ref {figr}(a), from approximately the 90th trading day (late April/early May) to approximately the 190th trading day (end of August/beginning of September), our transaction agent increased its holdings of `APPL' stocks and maintained the highest position. We looked at the share prices of each stock on 1 May 2020 versus 1 September 2020 and calculated the return on each stock for that period. The `APPL' stock had a return of 86.51$\%$ during this period, which is the highest among the stocks in the portfolio and much higher than the average return of 19.38$\%$ of the stocks in the portfolio. Combined with the Figure \ref {figr}(b), we can see that a large increase in holdings of `APPL' led to a significant increase in our agent's investment returns, and when it experienced shocks and declines later in the year, our agent reduced a large portion of those holdings, allowing for the majority of the returns to be retained. These show that our portfolio management strategy has been successful in giving higher weights to stocks with good return capacity in the position vector and succeeded in stopping losses promptly in a declining market without relying on stock price forecasts. 

\begin{figure}
	\centering
	\includegraphics[width=8cm]{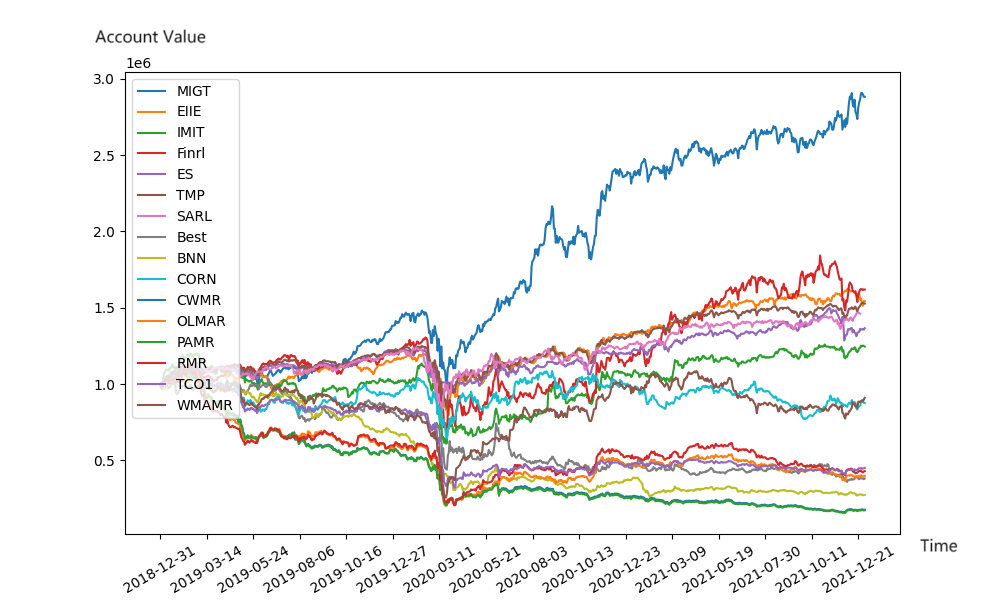}
	\caption{Results of the comparative experiments of data from 2019-2021.}\label{figa}
\end{figure}

To validate MIGT's performance over the long term, we use the 2019-2021 three-year data, i.e., combining dataset 1, 2, and 3, as a comparative experiment for the test set. As given in table \ref {tbla}, MIGT has at least 126.21$\%$ more cumulative returns than the comparative strategies. This suggests a greater advantage in the long-term return capacity of our strategy. As we can see from Figure \ref {figa}, MIGT begins to show a large advantage (about 50$\%$) in the early-middle period, and this advantage grows larger and larger (over 100$\%$) over time.

\begin{table*}
        \centering
	\caption{Three years of data from 2019-2021 were used as a test set to compare the results of the experiments, with the best results for each metric given in bold.}\label{tbla}
	
	\begin{tabular}{l l l l l}
		\toprule
		Strategies& Cumulative returns & Sharpe ratio & Omega ratio & Sortino ratio \\  
		\midrule
		\pmb{MIGT} & \pmb{1.88143} & \pmb{1.29061} & \pmb{1.28029} & \pmb{2.06325} \\
        EIIE & 0.55154 & 0.60861 & 1.14241 & 1.05018 \\
        IMIT  & 0.23495 &	0.28639 &	1.06158 &	0.41472 \\
        Finrl & 0.61929 & 0.52287 & 1.11752 & 0.92588 \\
        ES & 0.40192 & 0.57728 & 1.11972 & 1.03286 \\
        TMP & 0.53258 &	0.62228 &	1.14459 &	0.87164 \\
        SARL & 0.46130 &	0.54132 &	1.12693 &	0.75890 \\
        BEST & -0.61385 & -0.63775 & 0.88001 & -0.80744 \\
        BNN & -0.72627 & -1.25120 & 0.78986 & -1.49930 \\
        CORN & -0.12466 & -0.07156 & 0.98656 & 0.03114 \\
        CWMR & -0.82279 & -1.29828 & 0.75522 & -1.58723 \\
        OLMAR & -0.60072 & -0.53739 & 0.89345 & -0.61732 \\
        PAMR & -0.82657 & -1.31035 & 0.75346 & -1.59889 \\
        RMR & -0.57034 & -0.51821 & 0.89802 & -0.59114 \\
        TCO1 & -0.54912 & -0.63553 & 0.86083 & -0.73330 \\
        WMAMR & -0.08868 & 0.07072 & 1.01535 & 0.19846 \\
		\bottomrule
	\end{tabular}
\end{table*}

\subsection{Ablation Study}\label{}

Section \ref {Results} illustrates the performance of the MIGT framework. To analyze the impact of the individual modules on the effectiveness of our framework, we perform ablation experiments. The experiments provide a principled approach to understanding how and why the proposed framework works. To investigate the effectiveness of the Gated Instance Attention module, we modify it for the following version branches: the version with instance normalization removed (MIGT\_w/o\_Norm), the version with the gating layer removed (MIGT\_w/o\_Gating), and the version with the Transformer variant removed (MIGT\_w/o\_Transformer).

\begin{table*}
        \centering
	\caption{Results of the ablation experiments, where the best results for each metric are in bold.}\label{tbl3}
        \begin{tabular}{l l l l l}
		\toprule
		Strategies& Cumulative returns & Sharpe ratio & Omega ratio & Sortino ratio \\  
		\toprule
		Dataset 1\\
		\midrule
		\pmb{MIGT}  &\pmb{0.38443}  & \pmb{1.72498}  &  \pmb{1.33987} &  \pmb{2.74069}       \\
		MIGT\_w/o\_Norm   & 0.30139            & 1.46608       & 1.28920     & 2.39840        \\
		MIGT\_w/o\_Gating   & 0.24648            & 1.46140      & 1.27858     & 2.39581       \\
		MIGT\_w/o\_Transformer & 0.14537            & 0.73047      & 1.12926     & 1.29738       \\
		\toprule
		Dataset 2\\  
		\midrule
		\pmb{MIGT}     & \pmb{0.24440}      & \pmb{0.70689}      & \pmb{1.14642}     & \pmb{1.18089}       \\
		MIGT\_w/o\_Norm   & 0.22806 &	0.59541	& 1.12202	& 0.93648        \\
		MIGT\_w/o\_Gating   & 0.20915 &	0.59697	& 1.12509	& 0.93752       \\
		MIGT\_w/o\_Transformer & 0.06440 &	0.30429	& 1.05949	& 0.55714       \\
		\toprule
		Dataset 3\\  
		\midrule
		\pmb{MIGT}        & \pmb{0.28336} &	\pmb{1.30133}	& \pmb{1.24129}	& \pmb{2.18922}       \\
		MIGT\_w/o\_Norm   & 0.26727 &	1.19381	& 1.22241	& 1.93993       \\
		MIGT\_w/o\_Gating   & 0.25453 &	1.25659	& 1.22870	& 2.10863       \\
		MIGT\_w/o\_Transformer & 0.07674 &	0.34320	& 1.05727	& 0.74882       \\
				\bottomrule
	\end{tabular}
\end{table*}

\subsubsection{Overall ablation experiments with cumulative returns}\label{}
The results of the ablation experiments in Table \ref {tbl3} show that the return and ratio indicators for each group that underwent ablation are worse than MIGT. MIGT\_w/o\_Norm's impact on cumulative returns averaged -5.79$\%$, which suggests that the use of instance normalization to mitigate the impact of outliers has an obvious effect. However, this gap narrows to 1.61$\%$ in the experiments with 2021 data, suggesting that the outliers all have different degrees of impact on return capacity under different data sets. MIGT\_w/o\_Gating has a higher impact (-8.67$\%$), where the cumulative return in the experiment for the dataset 1 was 30.14$\%$, the difference with the full framework amounted to -8.39$\%$. This suggests that improving the stability of DRL training enhances the portfolio strategy's profitability. MIGT\_w/o\_Transformer model reduced both the return and the risk-return ratio to the lowest levels in the ablation experiment. The three-year average annual rate of return decreased by 22.76$\%$. The strategy applying Attention and Memory Trajectory increases the return by an average of 5.25$\%$ compared to the basic reinforcement learning framework. This shows that using the Gated Instance Attention module with a Transformer variant to handle stock data adequately has better performance.

\subsubsection{Ablation experiments for instance normalization}\label{}
\begin{figure*}
	\centering
	\includegraphics[width=15cm]{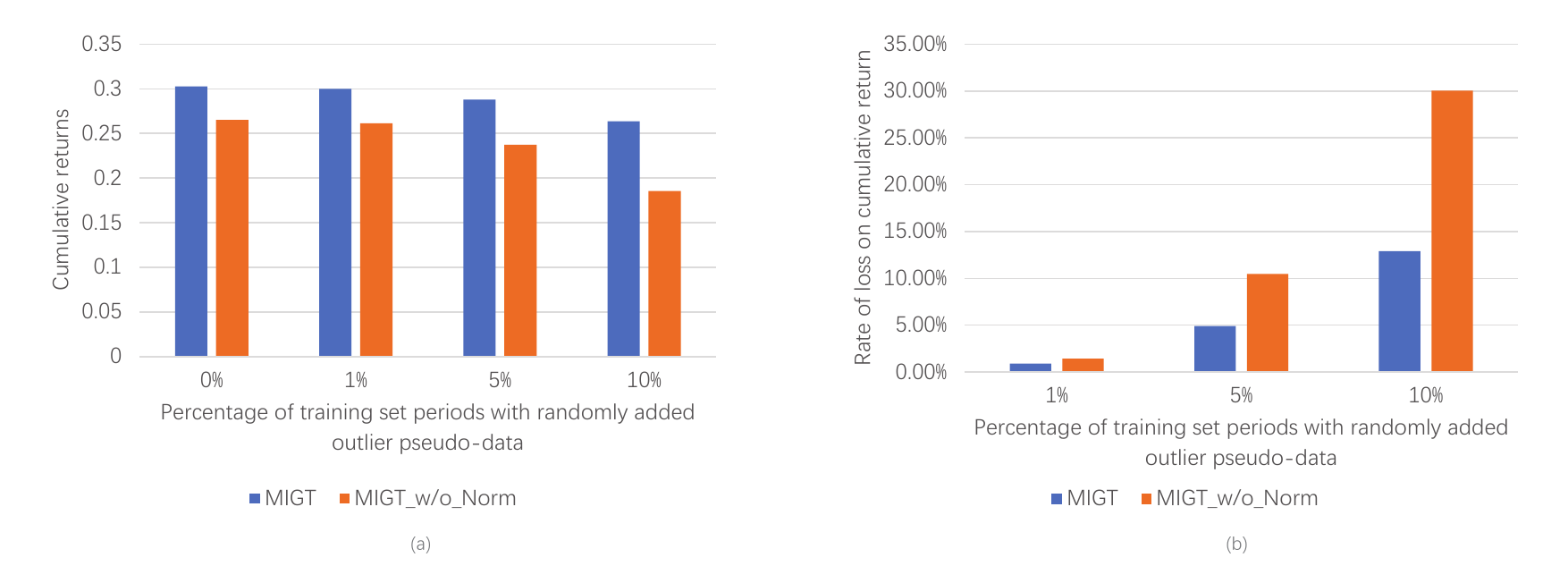}
	\caption{The effect of adding outlier pseudo-data on the cumulative return capacity of the strategy. Figure (a) shows the average cumulative return and Figure (b) shows the percentage of return lost compared to when no pseudo-data is added.}\label{figab1}
\end{figure*}

We introduce pseudo-data into the training dataset to test the effectiveness of instance normalization in mitigating the impact of outliers. Figure \ref {figab1} shows that when pseudo-data is not added, the average cumulative return of the MIGT strategy with instance normalization is 0.3027, which is higher than that without 0.2652. This shows the effect of instance normalization itself on the raw data. As the percentage of pseudo-data added increases, the average cumulative returns of both strategies decrease. This indicates that the outliers do have a negative impact on strategy performance, but the average cumulative return of the MIGT strategy using normalization decreases to a lesser extent. When 5$\%$ pseudo-data is added, the MIGT strategy loses 4.91$\%$ of its return and the MIGT\_w/o\_Norm strategy loses up to 22.67$\%$. When 10$\%$ pseudo-data is added, the MIGT strategy return only decreases to 0.2512 with 17.01$\%$ loss, while the unused one decreases to 0.1390 with 47.60$\%$ loss. This suggests that using instance normalization can significantly mitigate the negative impact of outliers and help the strategy maintain higher and more stable cumulative returns.

\subsubsection{Ablation experiments for LGU Gating Layer}\label{}
\begin{figure*}
	\centering
	\includegraphics[width=15cm]{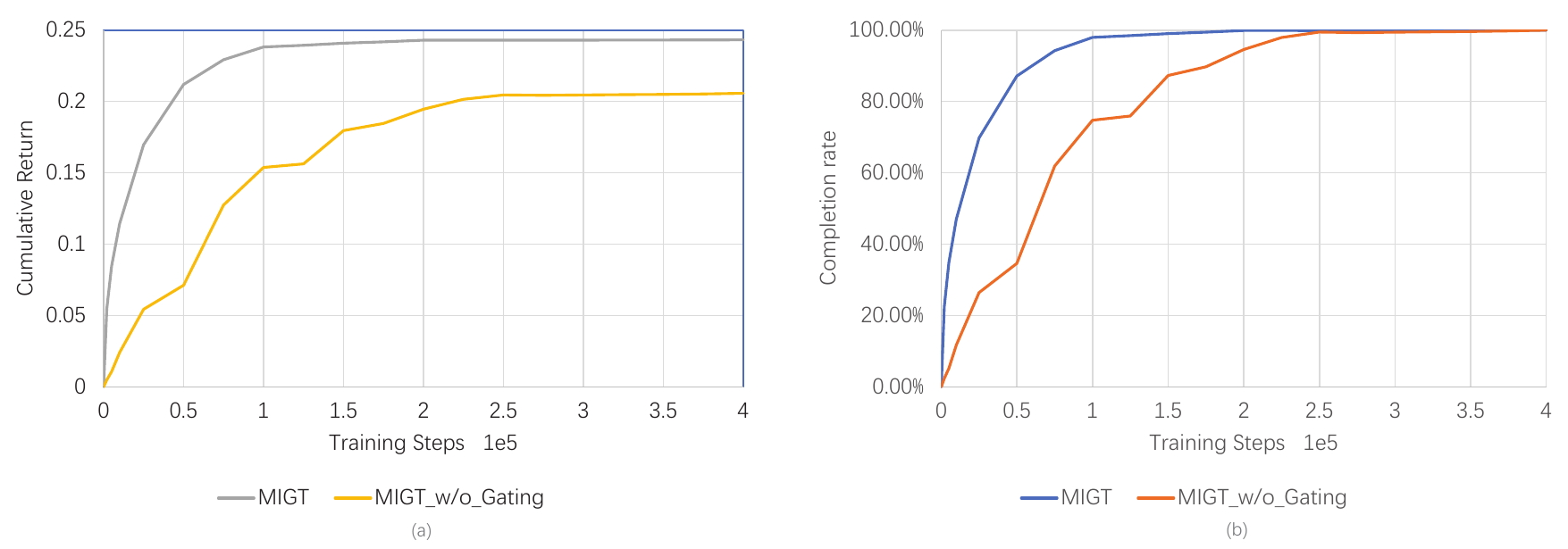}
	\caption{Line plots of training steps and cumulative returns for DRL portfolio management strategies. Figure (a) is by average cumulative return. Figure (b) is the ratio of cumulative return to this number at the completion of training.}\label{figab2}
\end{figure*}

We conducted ablation experiments, as given in Figure \ref{figab2}, to determine if the LGU Gating Layer improves the training of DRL. The cumulative return of the MIGT strategy increases steadily with the increase of training steps, reaching a peak of 0.238166 at 10,000 steps, and then remains stable at about 0.2429. The cumulative return of the MIGT\_w/o\_Gating strategy also rises with the increase of training steps but with a much smaller increase. The cumulative return of the MIGT strategy is significantly higher than that of the MIGT\_w/o\_Gating strategy, especially at the early stage when the number of training steps is small, and the difference between the two strategies is even larger. This illustrates the effect of using the LGU Gating Layer in the MIGT strategy. At the later stage, when there are more training steps, the cumulative returns of the two strategies stabilize, and the growth slows down, indicating that the strategies have basically converged. However, MIGT can converge to a higher level of cumulative returns than the strategy without a gating layer, suggesting that using the LGU gating layer results in a faster and more efficient output of the strategy.

\section{Conclusions and Future Work}\label{s6}

In this work, we have introduced MIGT, a novel framework for portfolio management using DRL, which enhances the performance of the portfolio framework with Gated Instance Attention, memory Trajectory, and MLP. The proposed Gated Instance Attention module combines a variant of Transformer, instance normalization, and LGU gating layer, which is crucial for increased profitability, and improved convergence for effective strategies. Our framework overperforms previous reinforcement learning portfolio frameworks in handling multi-dimensional data by resulting in higher stability of reinforcement learning, less influence of outliers, and a larger range of time series data processing. Experiments show the superiority of MIGT, as demonstrated by achieving higher return and risk-return metrics.

However, our current work has the following limitations. First, while our MIGT strategy achieves higher returns compared to other strategies, its advantage is less significant in terms of the Omega ratio which accounts for risk-adjusted returns. Our portfolio management framework utilizes some risk-related factors as input, but more comprehensive risk modeling could further improve risk-adjusted performance \citep{Maeso2020Maximizing}. Second, the Gated Instance Attention mechanism enhances our strategy's performance, but the network architecture and reinforcement learning algorithms can be further optimized for stock data and stock market characteristics. Third, short-selling operation is not allowed in our research assumptions. However, it provides valuable hedging and profit opportunities in practice, especially during market downturns \citep{Wang2020Shorting, Umar2021Media}. The constrained long-only formulation limits the strategy space and practical applicability in real-world portfolio management involving shorts. 

For future research, we suggest the following directions. First, we expand the risk modeling capabilities of our framework by incorporating additional risk factors as inputs. A more comprehensive set of risk metrics could further improve the risk-adjusted returns of our portfolio strategies. Second, we continue to optimize the neural architecture, with a focus on the attention components. By customizing the attention design for stock data properties, such as periodicity and volatility clustering, we can potentially improve performance. We will explore modifications to the Scaled Dot-Product Attention and other attentive mechanisms. Third, we enable flexible mixed long/short positions to allow our reinforcement learning agents to learn more advanced strategies with improved performance in both bull and bear regimes. At the same time, risk control mechanisms need to be in place to prevent massive losses from overleveraged short positions. For example, the reinforcement learning formulation should include additional guardrails and constraints around shorting as part of the action space. 

\section*{Declarations}

\pmb{Funding} No funding was received for this work.\\
\\
\pmb{Data Availability} Data will be made available on request.\\
\\
\pmb{Conflict of interest} All authors declare that he has no conflict of interest.\\
\\
\pmb{Ethical approval} This article does not contain any studies with human participants or animals performed by any of the authors.\\
\\
\pmb{Informed consent} Informed consent was obtained from all individual participants included in the study


\bibliography{references}

\begin{thebibliography}{}
\renewcommand{\doi}[1]{\url{https://doi.org/#1}}
\bibcommenthead

\bibitem [\protect \citeauthoryear {%
Ale%
, Hartford%
\BCBL {}\ \BBA {} Slater%
}{%
Ale%
\ \protect \BOthers {.}}{%
{\protect \APACyear {2020}}%
}]{%
Ale2020Dragons}
\APACinsertmetastar {%
Ale2020Dragons}%
\begin{APACrefauthors}%
Ale, B.%
, Hartford, D.%
\BCBL {} Slater, D.H.%
\end{APACrefauthors}%
\unskip\
\newblock
\APACrefYearMonthDay{2020}{}{}.
\newblock
{\BBOQ}\APACrefatitle {Dragons, black swans and decisions.} {Dragons, black swans and decisions.}{\BBCQ}
\newblock
\APACjournalVolNumPages{Environmental research}{183}{}{109127,}
\newblock
\begin{APACrefDOI} \doi{10.1016/j.envres.2020.109127} \end{APACrefDOI}
\newblock

\newblock

\PrintBackRefs{\CurrentBib}

\bibitem [\protect \citeauthoryear {%
Altan%
\ \BBA {} Karasu%
}{%
Altan%
\ \BBA {} Karasu%
}{%
{\protect \APACyear {2022}}%
}]{%
Altan2022}
\APACinsertmetastar {%
Altan2022}%
\begin{APACrefauthors}%
Altan, A.%
\BCBT {}\ \BBA {} Karasu, S.%
\end{APACrefauthors}%
\unskip\
\newblock
\APACrefYearMonthDay{2022}{}{}.
\newblock
{\BBOQ}\APACrefatitle {Crude oil time series prediction model based on LSTM network with chaotic Henry gas solubility optimization} {Crude oil time series prediction model based on lstm network with chaotic henry gas solubility optimization}.{\BBCQ}
\newblock
\APACjournalVolNumPages{Energy}{242}{}{,}
\newblock
\begin{APACrefDOI} \doi{10.1016/j.energy.2021.122964} \end{APACrefDOI}
\newblock

\newblock

\PrintBackRefs{\CurrentBib}

\bibitem [\protect \citeauthoryear {%
Arulkumaran%
, Deisenroth%
, Brundage%
\BCBL {}\ \BBA {} Bharath%
}{%
Arulkumaran%
\ \protect \BOthers {.}}{%
{\protect \APACyear {2017}}%
}]{%
Arulkumaran2017}
\APACinsertmetastar {%
Arulkumaran2017}%
\begin{APACrefauthors}%
Arulkumaran, K.%
, Deisenroth, M.P.%
, Brundage, M.%
\BCBL {} Bharath, A.A.%
\end{APACrefauthors}%
\unskip\
\newblock
\APACrefYearMonthDay{2017}{}{}.
\newblock
\APACrefbtitle {Deep reinforcement learning: A brief survey} {Deep reinforcement learning: A brief survey}\ (\BVOL~34).
\PrintBackRefs{\CurrentBib}

\bibitem [\protect \citeauthoryear {%
Avramov%
}{%
Avramov%
}{%
{\protect \APACyear {2002}}%
}]{%
Avramov2002}
\APACinsertmetastar {%
Avramov2002}%
\begin{APACrefauthors}%
Avramov, D.%
\end{APACrefauthors}%
\unskip\
\newblock
\APACrefYearMonthDay{2002}{}{}.
\newblock
{\BBOQ}\APACrefatitle {Stock return predictability and model uncertainty} {Stock return predictability and model uncertainty}.{\BBCQ}
\newblock
\APACjournalVolNumPages{Journal of Financial Economics}{64}{}{,}
\newblock
\begin{APACrefDOI} \doi{10.1016/S0304-405X(02)00131-9} \end{APACrefDOI}
\newblock

\newblock

\PrintBackRefs{\CurrentBib}

\bibitem [\protect \citeauthoryear {%
Baxter%
\ \BBA {} Puterman%
}{%
Baxter%
\ \BBA {} Puterman%
}{%
{\protect \APACyear {1995}}%
}]{%
Baxter1995}
\APACinsertmetastar {%
Baxter1995}%
\begin{APACrefauthors}%
Baxter, L.A.%
\BCBT {}\ \BBA {} Puterman, M.L.%
\end{APACrefauthors}%
\unskip\
\newblock
\APACrefYearMonthDay{1995}{}{}.
\newblock
{\BBOQ}\APACrefatitle {Markov Decision Processes: Discrete Stochastic Dynamic Programming} {Markov decision processes: Discrete stochastic dynamic programming}.{\BBCQ}
\newblock
\APACjournalVolNumPages{Technometrics}{37}{}{,}
\newblock
\begin{APACrefDOI} \doi{10.2307/1269932} \end{APACrefDOI}
\newblock

\newblock

\PrintBackRefs{\CurrentBib}

\bibitem [\protect \citeauthoryear {%
Bello%
, Zoph%
, Vasudevan%
\BCBL {}\ \BBA {} Le%
}{%
Bello%
\ \protect \BOthers {.}}{%
{\protect \APACyear {2016}}%
}]{%
2016Neural}
\APACinsertmetastar {%
2016Neural}%
\begin{APACrefauthors}%
Bello, I.%
, Zoph, B.%
, Vasudevan, V.%
\BCBL {} Le, Q.V.%
\end{APACrefauthors}%
\unskip\
\newblock
\APACrefYearMonthDay{2016}{}{}.
\newblock
{\BBOQ}\APACrefatitle {Neural Optimizer Search with Reinforcement Learning} {Neural optimizer search with reinforcement learning}.{\BBCQ}
\newblock

\newblock

\newblock

\PrintBackRefs{\CurrentBib}

\bibitem [\protect \citeauthoryear {%
Bengio%
\ \BBA {} Courville%
}{%
Bengio%
\ \BBA {} Courville%
}{%
{\protect \APACyear {2013}}%
}]{%
2013Deep}
\APACinsertmetastar {%
2013Deep}%
\begin{APACrefauthors}%
Bengio, Y.%
\BCBT {}\ \BBA {} Courville, A.C.%
\end{APACrefauthors}%
\unskip\
\newblock
\APACrefYearMonthDay{2013}{}{}.
\newblock
{\BBOQ}\APACrefatitle {Deep Learning of Representations} {Deep learning of representations}.{\BBCQ}
\newblock
 \APACrefbtitle {International Conference on Neural Information Processing.} {International conference on neural information processing.}
\PrintBackRefs{\CurrentBib}

\bibitem [\protect \citeauthoryear {%
Bollerslev%
, Li%
\BCBL {}\ \BBA {} Zhao%
}{%
Bollerslev%
\ \protect \BOthers {.}}{%
{\protect \APACyear {2020}}%
}]{%
Bollerslev2020Good}
\APACinsertmetastar {%
Bollerslev2020Good}%
\begin{APACrefauthors}%
Bollerslev, T.%
, Li, S.%
\BCBL {} Zhao, B.%
\end{APACrefauthors}%
\unskip\
\newblock
\APACrefYearMonthDay{2020}{}{}.
\newblock
{\BBOQ}\APACrefatitle {Good Volatility, Bad Volatility, and the Cross Section of Stock Returns} {Good volatility, bad volatility, and the cross section of stock returns}.{\BBCQ}
\newblock
\APACjournalVolNumPages{Journal of Financial and Quantitative Analysis}{55}{}{751 - 781,}
\newblock
\begin{APACrefDOI} \doi{10.1017/S0022109019000097} \end{APACrefDOI}
\newblock

\newblock

\PrintBackRefs{\CurrentBib}

\bibitem [\protect \citeauthoryear {%
Byrne%
\ \BBA {} Sakemoto%
}{%
Byrne%
\ \BBA {} Sakemoto%
}{%
{\protect \APACyear {2021}}%
}]{%
Byrne2021The}
\APACinsertmetastar {%
Byrne2021The}%
\begin{APACrefauthors}%
Byrne, J.P.%
\BCBT {}\ \BBA {} Sakemoto, R.%
\end{APACrefauthors}%
\unskip\
\newblock
\APACrefYearMonthDay{2021}{}{}.
\newblock
{\BBOQ}\APACrefatitle {The conditional volatility premium on currency portfolios} {The conditional volatility premium on currency portfolios}.{\BBCQ}
\newblock
\APACjournalVolNumPages{Journal of International Financial Markets, Institutions and Money}{74}{}{101415,}
\newblock
\begin{APACrefDOI} \doi{10.1016/J.INTFIN.2021.101415} \end{APACrefDOI}
\newblock

\newblock

\PrintBackRefs{\CurrentBib}

\bibitem [\protect \citeauthoryear {%
Cao%
, Li%
\BCBL {}\ \BBA {} Fair%
}{%
Cao%
\ \protect \BOthers {.}}{%
{\protect \APACyear {2019}}%
}]{%
Cao2019Deep}
\APACinsertmetastar {%
Cao2019Deep}%
\begin{APACrefauthors}%
Cao, C.%
, Li, D.%
\BCBL {} Fair, I.%
\end{APACrefauthors}%
\unskip\
\newblock
\APACrefYearMonthDay{2019}{}{}.
\newblock
{\BBOQ}\APACrefatitle {Deep Learning-Based Decoding of Constrained Sequence Codes} {Deep learning-based decoding of constrained sequence codes}.{\BBCQ}
\newblock
\APACjournalVolNumPages{IEEE Journal on Selected Areas in Communications}{37}{}{2532-2543,}
\newblock
\begin{APACrefDOI} \doi{10.1109/JSAC.2019.2933954} \end{APACrefDOI}
\newblock

\newblock

\PrintBackRefs{\CurrentBib}

\bibitem [\protect \citeauthoryear {%
CAVDAR%
\ \BBA {} AYDIN%
}{%
CAVDAR%
\ \BBA {} AYDIN%
}{%
{\protect \APACyear {2020}}%
}]{%
Seyma2020}
\APACinsertmetastar {%
Seyma2020}%
\begin{APACrefauthors}%
CAVDAR, S.C.%
\BCBT {}\ \BBA {} AYDIN, A.D.%
\end{APACrefauthors}%
\unskip\
\newblock
\APACrefYearMonthDay{2020}{}{}.
\newblock
{\BBOQ}\APACrefatitle {Hybrid Model Approach to the Complexity of Stock Trading Decisions in Turkey} {Hybrid model approach to the complexity of stock trading decisions in turkey}.{\BBCQ}
\newblock
\APACjournalVolNumPages{Journal of Asian Finance, Economics and Business}{7}{}{,}
\newblock
\begin{APACrefDOI} \doi{10.13106/jafeb.2020.vol7.no10.009} \end{APACrefDOI}
\newblock

\newblock

\PrintBackRefs{\CurrentBib}

\bibitem [\protect \citeauthoryear {%
Chang%
, Li%
\BCBL {}\ \BBA {} Zeng%
}{%
Chang%
\ \protect \BOthers {.}}{%
{\protect \APACyear {2019}}%
}]{%
Chang2019}
\APACinsertmetastar {%
Chang2019}%
\begin{APACrefauthors}%
Chang, V.%
, Li, T.%
\BCBL {} Zeng, Z.%
\end{APACrefauthors}%
\unskip\
\newblock
\APACrefYearMonthDay{2019}{}{}.
\newblock
{\BBOQ}\APACrefatitle {Towards an improved Adaboost algorithmic method for computational financial analysis} {Towards an improved adaboost algorithmic method for computational financial analysis}.{\BBCQ}
\newblock
\APACjournalVolNumPages{Journal of Parallel and Distributed Computing}{134}{}{,}
\newblock
\begin{APACrefDOI} \doi{10.1016/j.jpdc.2019.07.014} \end{APACrefDOI}
\newblock

\newblock

\PrintBackRefs{\CurrentBib}

\bibitem [\protect \citeauthoryear {%
Charpentier%
, Elie%
\BCBL {}\ \BBA {} Remlinger%
}{%
Charpentier%
\ \protect \BOthers {.}}{%
{\protect \APACyear {2020}}%
}]{%
Charpentier2020Reinforcement}
\APACinsertmetastar {%
Charpentier2020Reinforcement}%
\begin{APACrefauthors}%
Charpentier, A.%
, Elie, R.%
\BCBL {} Remlinger, C.%
\end{APACrefauthors}%
\unskip\
\newblock
\APACrefYearMonthDay{2020}{}{}.
\newblock
{\BBOQ}\APACrefatitle {Reinforcement Learning in Economics and Finance} {Reinforcement learning in economics and finance}.{\BBCQ}
\newblock
\APACjournalVolNumPages{ArXiv}{abs/2003.10014}{}{,}
\newblock
\begin{APACrefDOI} \doi{10.1007/S10614-021-10119-4} \end{APACrefDOI}
\newblock

\newblock

\PrintBackRefs{\CurrentBib}

\bibitem [\protect \citeauthoryear {%
Chasparis%
\ \BBA {} Shamma%
}{%
Chasparis%
\ \BBA {} Shamma%
}{%
{\protect \APACyear {2012}}%
}]{%
Chasparis2012Distributed}
\APACinsertmetastar {%
Chasparis2012Distributed}%
\begin{APACrefauthors}%
Chasparis, G.C.%
\BCBT {}\ \BBA {} Shamma, J.%
\end{APACrefauthors}%
\unskip\
\newblock
\APACrefYearMonthDay{2012}{}{}.
\newblock
{\BBOQ}\APACrefatitle {Distributed Dynamic Reinforcement of Efficient Outcomes in Multiagent Coordination and Network Formation} {Distributed dynamic reinforcement of efficient outcomes in multiagent coordination and network formation}.{\BBCQ}
\newblock
\APACjournalVolNumPages{Dynamic Games and Applications}{2}{}{18-50,}
\newblock
\begin{APACrefDOI} \doi{10.1007/s13235-011-0038-z} \end{APACrefDOI}
\newblock

\newblock

\PrintBackRefs{\CurrentBib}

\bibitem [\protect \citeauthoryear {%
Chung%
, Gulcehre%
, Cho%
\BCBL {}\ \BBA {} Bengio%
}{%
Chung%
\ \protect \BOthers {.}}{%
{\protect \APACyear {2014}}%
}]{%
Chung2014}
\APACinsertmetastar {%
Chung2014}%
\begin{APACrefauthors}%
Chung, J.%
, Gulcehre, C.%
, Cho, K.%
\BCBL {} Bengio, Y.%
\end{APACrefauthors}%
\unskip\
\newblock
\APACrefYearMonthDay{2014}{}{}.
\newblock
{\BBOQ}\APACrefatitle {Empirical evaluation of gated recurrent neural networks on sequence modeling} {Empirical evaluation of gated recurrent neural networks on sequence modeling}.{\BBCQ}
\newblock
 \APACrefbtitle {NIPS 2014 Workshop on Deep Learning, December 2014.} {Nips 2014 workshop on deep learning, december 2014.}
\PrintBackRefs{\CurrentBib}

\bibitem [\protect \citeauthoryear {%
Cui%
, Sun%
\BCBL {}\ \BBA {} Su%
}{%
Cui%
\ \protect \BOthers {.}}{%
{\protect \APACyear {2022}}%
}]{%
Cui2022}
\APACinsertmetastar {%
Cui2022}%
\begin{APACrefauthors}%
Cui, B.%
, Sun, R.%
\BCBL {} Su, J.%
\end{APACrefauthors}%
\unskip\
\newblock
\APACrefYearMonthDay{2022}{}{}.
\newblock
{\BBOQ}\APACrefatitle {A Novel Deep Reinforcement Learning Strategy in Financial Portfolio Management} {A novel deep reinforcement learning strategy in financial portfolio management}.{\BBCQ}.
\PrintBackRefs{\CurrentBib}

\bibitem [\protect \citeauthoryear {%
Dai%
\ \protect \BOthers {.}}{%
Dai%
\ \protect \BOthers {.}}{%
{\protect \APACyear {2020}}%
}]{%
Dai2020}
\APACinsertmetastar {%
Dai2020}%
\begin{APACrefauthors}%
Dai, Z.%
, Yang, Z.%
, Yang, Y.%
, Carbonell, J.%
, Le, Q.V.%
\BCBL {} Salakhutdinov, R.%
\end{APACrefauthors}%
\unskip\
\newblock
\APACrefYearMonthDay{2020}{}{}.
\newblock
{\BBOQ}\APACrefatitle {Transformer-XL: Attentive language models beyond a fixed-length context} {Transformer-xl: Attentive language models beyond a fixed-length context}.{\BBCQ}.
\PrintBackRefs{\CurrentBib}

\bibitem [\protect \citeauthoryear {%
Dayan%
\ \BBA {} Niv%
}{%
Dayan%
\ \BBA {} Niv%
}{%
{\protect \APACyear {2008}}%
}]{%
Dayan2008Reinforcement}
\APACinsertmetastar {%
Dayan2008Reinforcement}%
\begin{APACrefauthors}%
Dayan, P.%
\BCBT {}\ \BBA {} Niv, Y.%
\end{APACrefauthors}%
\unskip\
\newblock
\APACrefYearMonthDay{2008}{}{}.
\newblock
{\BBOQ}\APACrefatitle {Reinforcement learning: The Good, The Bad and The Ugly} {Reinforcement learning: The good, the bad and the ugly}.{\BBCQ}
\newblock
\APACjournalVolNumPages{Current Opinion in Neurobiology}{18}{}{185-196,}
\newblock
\begin{APACrefDOI} \doi{10.1016/j.conb.2008.08.003} \end{APACrefDOI}
\newblock

\newblock

\PrintBackRefs{\CurrentBib}

\bibitem [\protect \citeauthoryear {%
\APACcitebtitle {Dow Jones Industrial Average}}{%
\APACcitebtitle {Dow Jones Industrial Average}}{%
{\protect \APACyear {2023}}%
}]{%
DJIA}
\APACinsertmetastar {%
DJIA}%
\APACrefbtitle {Dow Jones Industrial Average} {Dow jones industrial average}.
\newblock
\APACrefYear{2023}.
\newblock
\begin{APACrefURL} {https://us.spindices.com/indices/equity/dow-jones-industrial-average} \end{APACrefURL}
\PrintBackRefs{\CurrentBib}

\bibitem [\protect \citeauthoryear {%
Fil%
\ \BBA {} Kristoufek%
}{%
Fil%
\ \BBA {} Kristoufek%
}{%
{\protect \APACyear {2020}}%
}]{%
Fil2020Pairs}
\APACinsertmetastar {%
Fil2020Pairs}%
\begin{APACrefauthors}%
Fil, M.%
\BCBT {}\ \BBA {} Kristoufek, L.%
\end{APACrefauthors}%
\unskip\
\newblock
\APACrefYearMonthDay{2020}{}{}.
\newblock
{\BBOQ}\APACrefatitle {Pairs Trading in Cryptocurrency Markets} {Pairs trading in cryptocurrency markets}.{\BBCQ}
\newblock
\APACjournalVolNumPages{IEEE Access}{8}{}{172644-172651,}
\newblock
\begin{APACrefDOI} \doi{10.1109/ACCESS.2020.3024619} \end{APACrefDOI}
\newblock

\newblock

\PrintBackRefs{\CurrentBib}

\bibitem [\protect \citeauthoryear {%
Fousekis%
\ \BBA {} Tzaferi%
}{%
Fousekis%
\ \BBA {} Tzaferi%
}{%
{\protect \APACyear {2021}}%
}]{%
Fousekis2021Returns}
\APACinsertmetastar {%
Fousekis2021Returns}%
\begin{APACrefauthors}%
Fousekis, P.%
\BCBT {}\ \BBA {} Tzaferi, D.%
\end{APACrefauthors}%
\unskip\
\newblock
\APACrefYearMonthDay{2021}{}{}.
\newblock
{\BBOQ}\APACrefatitle {Returns and volume: Frequency connectedness in cryptocurrency markets} {Returns and volume: Frequency connectedness in cryptocurrency markets}.{\BBCQ}
\newblock
\APACjournalVolNumPages{Economic Modelling}{95}{}{13-20,}
\newblock
\begin{APACrefDOI} \doi{10.1016/j.econmod.2020.11.013} \end{APACrefDOI}
\newblock

\newblock

\PrintBackRefs{\CurrentBib}

\bibitem [\protect \citeauthoryear {%
L.~Gao%
\ \BBA {} Zhang%
}{%
L.~Gao%
\ \BBA {} Zhang%
}{%
{\protect \APACyear {2013}}%
}]{%
Gao2013}
\APACinsertmetastar {%
Gao2013}%
\begin{APACrefauthors}%
Gao, L.%
\BCBT {}\ \BBA {} Zhang, W.%
\end{APACrefauthors}%
\unskip\
\newblock
\APACrefYearMonthDay{2013}{}{}.
\newblock
{\BBOQ}\APACrefatitle {Weighted moving average passive aggressive algorithm for online portfolio selection} {Weighted moving average passive aggressive algorithm for online portfolio selection}.{\BBCQ}
\newblock
 (\BVOL~1).
\PrintBackRefs{\CurrentBib}

\bibitem [\protect \citeauthoryear {%
R.~Gao%
\ \protect \BOthers {.}}{%
R.~Gao%
\ \protect \BOthers {.}}{%
{\protect \APACyear {2022}}%
}]{%
Gao2022}
\APACinsertmetastar {%
Gao2022}%
\begin{APACrefauthors}%
Gao, R.%
, Gu, F.%
, Sun, R.%
, Stefanidis, A.%
, Ren, X.%
\BCBL {} Su, J.%
\end{APACrefauthors}%
\unskip\
\newblock
\APACrefYearMonthDay{2022}{}{}.
\newblock
{\BBOQ}\APACrefatitle {A Novel DenseNet-based Deep Reinforcement Framework for Portfolio Management} {A novel densenet-based deep reinforcement framework for portfolio management}.{\BBCQ}
\newblock
 (\BPG~158-165).
\PrintBackRefs{\CurrentBib}

\bibitem [\protect \citeauthoryear {%
Z.~Gao%
, Gao%
, Hu%
, Jiang%
\BCBL {}\ \BBA {} Su%
}{%
Z.~Gao%
\ \protect \BOthers {.}}{%
{\protect \APACyear {2020}}%
}]{%
Gao2020}
\APACinsertmetastar {%
Gao2020}%
\begin{APACrefauthors}%
Gao, Z.%
, Gao, Y.%
, Hu, Y.%
, Jiang, Z.%
\BCBL {} Su, J.%
\end{APACrefauthors}%
\unskip\
\newblock
\APACrefYearMonthDay{2020}{}{}.
\newblock
{\BBOQ}\APACrefatitle {Application of Deep Q-Network in Portfolio Management} {Application of deep q-network in portfolio management}.{\BBCQ}.
\PrintBackRefs{\CurrentBib}

\bibitem [\protect \citeauthoryear {%
Gershman%
, Blei%
\BCBL {}\ \BBA {} Niv%
}{%
Gershman%
\ \protect \BOthers {.}}{%
{\protect \APACyear {2010}}%
}]{%
Gershman2010Context}
\APACinsertmetastar {%
Gershman2010Context}%
\begin{APACrefauthors}%
Gershman, S.%
, Blei, D.%
\BCBL {} Niv, Y.%
\end{APACrefauthors}%
\unskip\
\newblock
\APACrefYearMonthDay{2010}{}{}.
\newblock
{\BBOQ}\APACrefatitle {Context, learning, and extinction.} {Context, learning, and extinction.}{\BBCQ}
\newblock
\APACjournalVolNumPages{Psychological review}{117 1}{}{197-209,}
\newblock
\begin{APACrefDOI} \doi{10.1037/a0017808} \end{APACrefDOI}
\newblock

\newblock

\PrintBackRefs{\CurrentBib}

\bibitem [\protect \citeauthoryear {%
Goodfellow%
\ \protect \BOthers {.}}{%
Goodfellow%
\ \protect \BOthers {.}}{%
{\protect \APACyear {2014}}%
}]{%
Goodfellow2014}
\APACinsertmetastar {%
Goodfellow2014}%
\begin{APACrefauthors}%
Goodfellow, I.J.%
, Pouget-Abadie, J.%
, Mirza, M.%
, Xu, B.%
, Warde-Farley, D.%
, Ozair, S.%
\BDBL {}Bengio, Y.%
\end{APACrefauthors}%
\unskip\
\newblock
\APACrefYearMonthDay{2014}{}{}.
\newblock
{\BBOQ}\APACrefatitle {Generative adversarial nets} {Generative adversarial nets}.{\BBCQ}
\newblock
 (\BVOL~3).
\PrintBackRefs{\CurrentBib}

\bibitem [\protect \citeauthoryear {%
Gu%
, Jiang%
\BCBL {}\ \BBA {} Su%
}{%
Gu%
\ \protect \BOthers {.}}{%
{\protect \APACyear {2021}}%
}]{%
Gu2021}
\APACinsertmetastar {%
Gu2021}%
\begin{APACrefauthors}%
Gu, F.%
, Jiang, Z.%
\BCBL {} Su, J.%
\end{APACrefauthors}%
\unskip\
\newblock
\APACrefYearMonthDay{2021}{}{}.
\newblock
{\BBOQ}\APACrefatitle {Application of Features and Neural Network to Enhance the Performance of Deep Reinforcement Learning in Portfolio Management} {Application of features and neural network to enhance the performance of deep reinforcement learning in portfolio management}.{\BBCQ}.
\PrintBackRefs{\CurrentBib}

\bibitem [\protect \citeauthoryear {%
Guan%
\ \BBA {} Liu%
}{%
Guan%
\ \BBA {} Liu%
}{%
{\protect \APACyear {2022}}%
}]{%
Guan2021}
\APACinsertmetastar {%
Guan2021}%
\begin{APACrefauthors}%
Guan, M.%
\BCBT {}\ \BBA {} Liu, X\BHBI Y.%
\end{APACrefauthors}%
\unskip\
\newblock
\APACrefYearMonthDay{2022}{}{}.
\newblock
{\BBOQ}\APACrefatitle {Explainable Deep Reinforcement Learning for Portfolio Management: An Empirical Approach} {Explainable deep reinforcement learning for portfolio management: An empirical approach}.{\BBCQ}
\newblock
 \APACrefbtitle {Proceedings of the Second ACM International Conference on AI in Finance.} {Proceedings of the second acm international conference on ai in finance.}
\newblock
\APACaddressPublisher{New York, NY, USA}{Association for Computing Machinery}.
\newblock
\begin{APACrefURL} {https://doi.org/10.1145/3490354.3494415} \end{APACrefURL}
\PrintBackRefs{\CurrentBib}

\bibitem [\protect \citeauthoryear {%
Gupta%
, Singal%
\BCBL {}\ \BBA {} Garg%
}{%
Gupta%
\ \protect \BOthers {.}}{%
{\protect \APACyear {2021}}%
}]{%
Gupta2021Deep}
\APACinsertmetastar {%
Gupta2021Deep}%
\begin{APACrefauthors}%
Gupta, S.%
, Singal, G.%
\BCBL {} Garg, D.%
\end{APACrefauthors}%
\unskip\
\newblock
\APACrefYearMonthDay{2021}{}{}.
\newblock
{\BBOQ}\APACrefatitle {Deep Reinforcement Learning Techniques in Diversified Domains: A Survey} {Deep reinforcement learning techniques in diversified domains: A survey}.{\BBCQ}
\newblock
\APACjournalVolNumPages{Archives of Computational Methods in Engineering}{28}{}{4715 - 4754,}
\newblock
\begin{APACrefDOI} \doi{10.1007/s11831-021-09552-3} \end{APACrefDOI}
\newblock

\newblock

\PrintBackRefs{\CurrentBib}

\bibitem [\protect \citeauthoryear {%
Györfi%
, Lugosi%
\BCBL {}\ \BBA {} Udina%
}{%
Györfi%
\ \protect \BOthers {.}}{%
{\protect \APACyear {2006}}%
}]{%
Laszlo2006}
\APACinsertmetastar {%
Laszlo2006}%
\begin{APACrefauthors}%
Györfi, L.%
, Lugosi, G.%
\BCBL {} Udina, F.%
\end{APACrefauthors}%
\unskip\
\newblock
\APACrefYearMonthDay{2006}{}{}.
\newblock
{\BBOQ}\APACrefatitle {Nonparametric kernel-based sequential investment strategies} {Nonparametric kernel-based sequential investment strategies}.{\BBCQ}
\newblock
\APACjournalVolNumPages{Mathematical Finance}{16}{}{,}
\newblock
\begin{APACrefDOI} \doi{10.1111/j.1467-9965.2006.00274.x} \end{APACrefDOI}
\newblock

\newblock

\PrintBackRefs{\CurrentBib}

\bibitem [\protect \citeauthoryear {%
Haarnoja%
, Zhou%
, Abbeel%
\BCBL {}\ \BBA {} Levine%
}{%
Haarnoja%
\ \protect \BOthers {.}}{%
{\protect \APACyear {2018}}%
}]{%
2018Soft}
\APACinsertmetastar {%
2018Soft}%
\begin{APACrefauthors}%
Haarnoja, T.%
, Zhou, A.%
, Abbeel, P.%
\BCBL {} Levine, S.%
\end{APACrefauthors}%
\unskip\
\newblock
\APACrefYearMonthDay{2018}{}{}.
\newblock
\APACrefbtitle {Soft Actor-Critic: Off-Policy Maximum Entropy Deep Reinforcement Learning with a Stochastic Actor.} {Soft actor-critic: Off-policy maximum entropy deep reinforcement learning with a stochastic actor.}
\PrintBackRefs{\CurrentBib}

\bibitem [\protect \citeauthoryear {%
Hu%
, Zhao%
\BCBL {}\ \BBA {} Khushi%
}{%
Hu%
\ \protect \BOthers {.}}{%
{\protect \APACyear {2021}}%
}]{%
Hu2021A}
\APACinsertmetastar {%
Hu2021A}%
\begin{APACrefauthors}%
Hu, Z.%
, Zhao, Y.%
\BCBL {} Khushi, M.%
\end{APACrefauthors}%
\unskip\
\newblock
\APACrefYearMonthDay{2021}{}{}.
\newblock
{\BBOQ}\APACrefatitle {A Survey of Forex and Stock Price Prediction Using Deep Learning} {A survey of forex and stock price prediction using deep learning}.{\BBCQ}
\newblock
\APACjournalVolNumPages{Applied System Innovation}{}{}{,}
\newblock
\begin{APACrefDOI} \doi{10.3390/asi4010009} \end{APACrefDOI}
\newblock

\newblock

\PrintBackRefs{\CurrentBib}

\bibitem [\protect \citeauthoryear {%
Huang%
, Zhou%
, Li%
, Hoi%
\BCBL {}\ \BBA {} Zhou%
}{%
Huang%
\ \protect \BOthers {.}}{%
{\protect \APACyear {2016}}%
}]{%
Huang2016}
\APACinsertmetastar {%
Huang2016}%
\begin{APACrefauthors}%
Huang, D.J.%
, Zhou, J.%
, Li, B.%
, Hoi, S.C.%
\BCBL {} Zhou, S.%
\end{APACrefauthors}%
\unskip\
\newblock
\APACrefYearMonthDay{2016}{}{}.
\newblock
{\BBOQ}\APACrefatitle {Robust Median Reversion Strategy for Online Portfolio Selection} {Robust median reversion strategy for online portfolio selection}.{\BBCQ}
\newblock
\APACjournalVolNumPages{IEEE Transactions on Knowledge and Data Engineering}{28}{}{,}
\newblock
\begin{APACrefDOI} \doi{10.1109/TKDE.2016.2563433} \end{APACrefDOI}
\newblock

\newblock

\PrintBackRefs{\CurrentBib}

\bibitem [\protect \citeauthoryear {%
Hung%
}{%
Hung%
}{%
{\protect \APACyear {2016}}%
}]{%
Hung2016}
\APACinsertmetastar {%
Hung2016}%
\begin{APACrefauthors}%
Hung, N.H.%
\end{APACrefauthors}%
\unskip\
\newblock
\APACrefYearMonthDay{2016}{}{}.
\newblock
{\BBOQ}\APACrefatitle {Various moving average convergence divergence trading strategies: A comparison} {Various moving average convergence divergence trading strategies: A comparison}.{\BBCQ}
\newblock
\APACjournalVolNumPages{Investment Management and Financial Innovations}{13}{}{,}
\newblock
\begin{APACrefDOI} \doi{10.21511/imfi.13(2-2).2016.11} \end{APACrefDOI}
\newblock

\newblock

\PrintBackRefs{\CurrentBib}

\bibitem [\protect \citeauthoryear {%
Jiang%
\ \BBA {} Liang%
}{%
Jiang%
\ \BBA {} Liang%
}{%
{\protect \APACyear {2018}}%
}]{%
Jiang2018}
\APACinsertmetastar {%
Jiang2018}%
\begin{APACrefauthors}%
Jiang, Z.%
\BCBT {}\ \BBA {} Liang, J.%
\end{APACrefauthors}%
\unskip\
\newblock
\APACrefYearMonthDay{2018}{}{}.
\newblock
{\BBOQ}\APACrefatitle {Cryptocurrency portfolio management with deep reinforcement learning} {Cryptocurrency portfolio management with deep reinforcement learning}.{\BBCQ}
\newblock
 (\BVOL\ 2018-January).
\PrintBackRefs{\CurrentBib}

\bibitem [\protect \citeauthoryear {%
Jiang%
, Xu%
\BCBL {}\ \BBA {} Liang%
}{%
Jiang%
\ \protect \BOthers {.}}{%
{\protect \APACyear {2017}}%
}]{%
Jiang2017}
\APACinsertmetastar {%
Jiang2017}%
\begin{APACrefauthors}%
Jiang, Z.%
, Xu, D.%
\BCBL {} Liang, J.%
\end{APACrefauthors}%
\unskip\
\newblock
\APACrefYearMonthDay{2017}{6}{}.
\newblock
{\BBOQ}\APACrefatitle {A Deep Reinforcement Learning Framework for the Financial Portfolio Management Problem} {A deep reinforcement learning framework for the financial portfolio management problem}.{\BBCQ}
\newblock

\newblock
\begin{APACrefDOI} \doi{10.48550} \end{APACrefDOI}
\newblock

\newblock

\PrintBackRefs{\CurrentBib}

\bibitem [\protect \citeauthoryear {%
Jurafsky%
\ \BBA {} Martin%
}{%
Jurafsky%
\ \BBA {} Martin%
}{%
{\protect \APACyear {2009}}%
}]{%
1214993}
\APACinsertmetastar {%
1214993}%
\begin{APACrefauthors}%
Jurafsky, D.%
\BCBT {}\ \BBA {} Martin, J.H.%
\end{APACrefauthors}%
\unskip\
\newblock
\APACrefYear{2009}.
\newblock
\APACrefbtitle {Speech and Language Processing (2nd Edition)} {Speech and language processing (2nd edition)}.
\newblock
\APACaddressPublisher{USA}{Prentice-Hall, Inc.}
\PrintBackRefs{\CurrentBib}

\bibitem [\protect \citeauthoryear {%
Kandanaarachchi%
, Muñoz%
, Hyndman%
\BCBL {}\ \BBA {} Smith-Miles%
}{%
Kandanaarachchi%
\ \protect \BOthers {.}}{%
{\protect \APACyear {2020}}%
}]{%
Kandanaarachchi2020}
\APACinsertmetastar {%
Kandanaarachchi2020}%
\begin{APACrefauthors}%
Kandanaarachchi, S.%
, Muñoz, M.A.%
, Hyndman, R.J.%
\BCBL {} Smith-Miles, K.%
\end{APACrefauthors}%
\unskip\
\newblock
\APACrefYearMonthDay{2020}{}{}.
\newblock
{\BBOQ}\APACrefatitle {On normalization and algorithm selection for unsupervised outlier detection} {On normalization and algorithm selection for unsupervised outlier detection}.{\BBCQ}
\newblock
\APACjournalVolNumPages{Data Mining and Knowledge Discovery}{34}{}{,}
\newblock
\begin{APACrefDOI} \doi{10.1007/s10618-019-00661-z} \end{APACrefDOI}
\newblock

\newblock

\PrintBackRefs{\CurrentBib}

\bibitem [\protect \citeauthoryear {%
Keating%
\ \BBA {} Shadwick%
}{%
Keating%
\ \BBA {} Shadwick%
}{%
{\protect \APACyear {2002}}%
}]{%
Keating2002}
\APACinsertmetastar {%
Keating2002}%
\begin{APACrefauthors}%
Keating, C.%
\BCBT {}\ \BBA {} Shadwick, W.F.%
\end{APACrefauthors}%
\unskip\
\newblock
\APACrefYearMonthDay{2002}{}{}.
\newblock
{\BBOQ}\APACrefatitle {An introduction to Omega} {An introduction to omega}.{\BBCQ}
\newblock
\APACjournalVolNumPages{The Finance Development Centre Ltd}{}{}{,}
\newblock

\newblock

\PrintBackRefs{\CurrentBib}

\bibitem [\protect \citeauthoryear {%
Khader%
, Yin%
, Falco%
\BCBL {}\ \BBA {} Kragic%
}{%
Khader%
\ \protect \BOthers {.}}{%
{\protect \APACyear {2021}}%
}]{%
Khader2021Learning}
\APACinsertmetastar {%
Khader2021Learning}%
\begin{APACrefauthors}%
Khader, S.A.%
, Yin, H.%
, Falco, P.%
\BCBL {} Kragic, D.%
\end{APACrefauthors}%
\unskip\
\newblock
\APACrefYearMonthDay{2021}{}{}.
\newblock
{\BBOQ}\APACrefatitle {Learning Deep Energy Shaping Policies for Stability-Guaranteed Manipulation} {Learning deep energy shaping policies for stability-guaranteed manipulation}.{\BBCQ}
\newblock
\APACjournalVolNumPages{IEEE Robotics and Automation Letters}{6}{}{8583-8590,}
\newblock
\begin{APACrefDOI} \doi{10.1109/LRA.2021.3111962} \end{APACrefDOI}
\newblock

\newblock

\PrintBackRefs{\CurrentBib}

\bibitem [\protect \citeauthoryear {%
Koziarski%
}{%
Koziarski%
}{%
{\protect \APACyear {2020}}%
}]{%
Koziarski2020}
\APACinsertmetastar {%
Koziarski2020}%
\begin{APACrefauthors}%
Koziarski, M.%
\end{APACrefauthors}%
\unskip\
\newblock
\APACrefYearMonthDay{2020}{}{}.
\newblock
{\BBOQ}\APACrefatitle {Radial-Based Undersampling for imbalanced data classification} {Radial-based undersampling for imbalanced data classification}.{\BBCQ}
\newblock
\APACjournalVolNumPages{Pattern Recognition}{102}{}{,}
\newblock
\begin{APACrefDOI} \doi{10.1016/j.patcog.2020.107262} \end{APACrefDOI}
\newblock

\newblock

\PrintBackRefs{\CurrentBib}

\bibitem [\protect \citeauthoryear {%
Kumar%
, Yadav%
, Gupta%
\BCBL {}\ \BBA {} Mehlawat%
}{%
Kumar%
\ \protect \BOthers {.}}{%
{\protect \APACyear {2021}}%
}]{%
Kumar2021A}
\APACinsertmetastar {%
Kumar2021A}%
\begin{APACrefauthors}%
Kumar, A.%
, Yadav, S.%
, Gupta, P.%
\BCBL {} Mehlawat, M.K.%
\end{APACrefauthors}%
\unskip\
\newblock
\APACrefYearMonthDay{2021}{}{}.
\newblock
{\BBOQ}\APACrefatitle {A Credibilistic Multiobjective Multiperiod Efficient Portfolio Selection Approach Using Data Envelopment Analysis} {A credibilistic multiobjective multiperiod efficient portfolio selection approach using data envelopment analysis}.{\BBCQ}
\newblock
\APACjournalVolNumPages{IEEE Transactions on Engineering Management}{PP}{}{1-15,}
\newblock
\begin{APACrefDOI} \doi{10.1109/TEM.2021.3072543} \end{APACrefDOI}
\newblock

\newblock

\PrintBackRefs{\CurrentBib}

\bibitem [\protect \citeauthoryear {%
Lecun%
, Bengio%
\BCBL {}\ \BBA {} Hinton%
}{%
Lecun%
\ \protect \BOthers {.}}{%
{\protect \APACyear {2015}}%
}]{%
2015Deep}
\APACinsertmetastar {%
2015Deep}%
\begin{APACrefauthors}%
Lecun, Y.%
, Bengio, Y.%
\BCBL {} Hinton, G.%
\end{APACrefauthors}%
\unskip\
\newblock
\APACrefYearMonthDay{2015}{}{}.
\newblock
{\BBOQ}\APACrefatitle {Deep learning} {Deep learning}.{\BBCQ}
\newblock
\APACjournalVolNumPages{Nature}{521}{7553}{436,}
\newblock

\newblock

\PrintBackRefs{\CurrentBib}

\bibitem [\protect \citeauthoryear {%
Lehnert%
, Littman%
\BCBL {}\ \BBA {} Frank%
}{%
Lehnert%
\ \protect \BOthers {.}}{%
{\protect \APACyear {2020}}%
}]{%
Lehnert2020Reward-predictive}
\APACinsertmetastar {%
Lehnert2020Reward-predictive}%
\begin{APACrefauthors}%
Lehnert, L.%
, Littman, M.L.%
\BCBL {} Frank, M.J.%
\end{APACrefauthors}%
\unskip\
\newblock
\APACrefYearMonthDay{2020}{}{}.
\newblock
{\BBOQ}\APACrefatitle {Reward-predictive representations generalize across tasks in reinforcement learning} {Reward-predictive representations generalize across tasks in reinforcement learning}.{\BBCQ}
\newblock
\APACjournalVolNumPages{PLoS Computational Biology}{16}{}{,}
\newblock
\begin{APACrefDOI} \doi{10.1371/journal.pcbi.1008317} \end{APACrefDOI}
\newblock

\newblock

\PrintBackRefs{\CurrentBib}

\bibitem [\protect \citeauthoryear {%
Li%
\ \BBA {} Hoi%
}{%
Li%
\ \BBA {} Hoi%
}{%
{\protect \APACyear {2012}}%
}]{%
Li2012}
\APACinsertmetastar {%
Li2012}%
\begin{APACrefauthors}%
Li, B.%
\BCBT {}\ \BBA {} Hoi, S.C.%
\end{APACrefauthors}%
\unskip\
\newblock
\APACrefYearMonthDay{2012}{}{}.
\newblock
{\BBOQ}\APACrefatitle {On-line portfolio selection with moving average reversion} {On-line portfolio selection with moving average reversion}.{\BBCQ}
\newblock
 (\BVOL~1).
\PrintBackRefs{\CurrentBib}

\bibitem [\protect \citeauthoryear {%
Li%
, Hoi%
\BCBL {}\ \BBA {} Gopalkrishnan%
}{%
Li%
, Hoi%
\BCBL {}\ \BBA {} Gopalkrishnan%
}{%
{\protect \APACyear {2011}}%
}]{%
Lii2011}
\APACinsertmetastar {%
Lii2011}%
\begin{APACrefauthors}%
Li, B.%
, Hoi, S.C.%
\BCBL {} Gopalkrishnan, V.%
\end{APACrefauthors}%
\unskip\
\newblock
\APACrefYearMonthDay{2011}{}{}.
\newblock
{\BBOQ}\APACrefatitle {CORN: Correlation-driven nonparametric learning approach for portfolio selection} {Corn: Correlation-driven nonparametric learning approach for portfolio selection}.{\BBCQ}
\newblock
\APACjournalVolNumPages{ACM Transactions on Intelligent Systems and Technology}{2}{}{,}
\newblock
\begin{APACrefDOI} \doi{10.1145/1961189.1961193} \end{APACrefDOI}
\newblock

\newblock

\PrintBackRefs{\CurrentBib}

\bibitem [\protect \citeauthoryear {%
Li%
, Hoi%
, Zhao%
\BCBL {}\ \BBA {} Gopalkrishnan%
}{%
Li%
, Hoi%
, Zhao%
\BCBL {}\ \BBA {} Gopalkrishnan%
}{%
{\protect \APACyear {2011}}%
}]{%
Li2011}
\APACinsertmetastar {%
Li2011}%
\begin{APACrefauthors}%
Li, B.%
, Hoi, S.C.%
, Zhao, P.%
\BCBL {} Gopalkrishnan, V.%
\end{APACrefauthors}%
\unskip\
\newblock
\APACrefYearMonthDay{2011}{}{}.
\newblock
{\BBOQ}\APACrefatitle {Confidence Weighted Mean Reversion strategy for on-line portfolio selection} {Confidence weighted mean reversion strategy for on-line portfolio selection}.{\BBCQ}
\newblock
 (\BVOL~15).
\PrintBackRefs{\CurrentBib}

\bibitem [\protect \citeauthoryear {%
Li%
, Wang%
, Huang%
\BCBL {}\ \BBA {} Hoi%
}{%
Li%
\ \protect \BOthers {.}}{%
{\protect \APACyear {2018}}%
}]{%
Li2018}
\APACinsertmetastar {%
Li2018}%
\begin{APACrefauthors}%
Li, B.%
, Wang, J.%
, Huang, D.%
\BCBL {} Hoi, S.C.%
\end{APACrefauthors}%
\unskip\
\newblock
\APACrefYearMonthDay{2018}{}{}.
\newblock
{\BBOQ}\APACrefatitle {Transaction cost optimization for online portfolio selection} {Transaction cost optimization for online portfolio selection}.{\BBCQ}
\newblock
\APACjournalVolNumPages{Quantitative Finance}{18}{}{,}
\newblock
\begin{APACrefDOI} \doi{10.1080/14697688.2017.1357831} \end{APACrefDOI}
\newblock

\newblock

\PrintBackRefs{\CurrentBib}

\bibitem [\protect \citeauthoryear {%
Li%
, Zhao%
, Hoi%
\BCBL {}\ \BBA {} Gopalkrishnan%
}{%
Li%
\ \protect \BOthers {.}}{%
{\protect \APACyear {2012}}%
}]{%
Lii2012}
\APACinsertmetastar {%
Lii2012}%
\begin{APACrefauthors}%
Li, B.%
, Zhao, P.%
, Hoi, S.C.%
\BCBL {} Gopalkrishnan, V.%
\end{APACrefauthors}%
\unskip\
\newblock
\APACrefYearMonthDay{2012}{}{}.
\newblock
{\BBOQ}\APACrefatitle {PAMR: Passive aggressive mean reversion strategy for portfolio selection} {Pamr: Passive aggressive mean reversion strategy for portfolio selection}.{\BBCQ}
\newblock
\APACjournalVolNumPages{Machine Learning}{87}{}{,}
\newblock
\begin{APACrefDOI} \doi{10.1007/s10994-012-5281-z} \end{APACrefDOI}
\newblock

\newblock

\PrintBackRefs{\CurrentBib}

\bibitem [\protect \citeauthoryear {%
C.~Liu%
, Ventre%
\BCBL {}\ \BBA {} Polukarov%
}{%
C.~Liu%
\ \protect \BOthers {.}}{%
{\protect \APACyear {2022}}%
}]{%
Liu101145}
\APACinsertmetastar {%
Liu101145}%
\begin{APACrefauthors}%
Liu, C.%
, Ventre, C.%
\BCBL {} Polukarov, M.%
\end{APACrefauthors}%
\unskip\
\newblock
\APACrefYearMonthDay{2022}{}{}.
\newblock
{\BBOQ}\APACrefatitle {Synthetic Data Augmentation for Deep Reinforcement Learning in Financial Trading} {Synthetic data augmentation for deep reinforcement learning in financial trading}.{\BBCQ}
\newblock
 \APACrefbtitle {Proceedings of the Third ACM International Conference on AI in Finance} {Proceedings of the third acm international conference on ai in finance}\ (\BPG~343–351).
\newblock
\APACaddressPublisher{New York, NY, USA}{Association for Computing Machinery}.
\newblock
\begin{APACrefURL} {https://doi.org/10.1145/3533271.3561704} \end{APACrefURL}
\PrintBackRefs{\CurrentBib}

\bibitem [\protect \citeauthoryear {%
X\BHBI Y.~Liu%
\ \protect \BOthers {.}}{%
X\BHBI Y.~Liu%
\ \protect \BOthers {.}}{%
{\protect \APACyear {2021}}%
}]{%
Liu2021}
\APACinsertmetastar {%
Liu2021}%
\begin{APACrefauthors}%
Liu, X\BHBI Y.%
, Yang, H.%
, Chen, Q.%
, Zhang, R.%
, Yang, L.%
, Xiao, B.%
\BCBL {} Wang, C.%
\end{APACrefauthors}%
\unskip\
\newblock
\APACrefYearMonthDay{2021}{}{}.
\newblock
{\BBOQ}\APACrefatitle {FinRL: A Deep Reinforcement Learning Library for Automated Stock Trading in Quantitative Finance} {Finrl: A deep reinforcement learning library for automated stock trading in quantitative finance}.{\BBCQ}
\newblock
\APACjournalVolNumPages{SSRN Electronic Journal}{}{}{,}
\newblock
\begin{APACrefDOI} \doi{10.2139/ssrn.3737859} \end{APACrefDOI}
\newblock

\newblock

\PrintBackRefs{\CurrentBib}

\bibitem [\protect \citeauthoryear {%
Z.T.Y.~Liu%
}{%
Z.T.Y.~Liu%
}{%
{\protect \APACyear {2018}}%
}]{%
2018Investor}
\APACinsertmetastar {%
2018Investor}%
\begin{APACrefauthors}%
Liu, Z.T.Y.%
\end{APACrefauthors}%
\unskip\
\newblock
\APACrefYearMonthDay{2018}{}{}.
\newblock
{\BBOQ}\APACrefatitle {Investor-Imitator: A Framework for Trading Knowledge Extraction} {Investor-imitator: A framework for trading knowledge extraction}.{\BBCQ}
\newblock
\APACjournalVolNumPages{SIGKDD explorations}{}{Udisk}{,}
\newblock

\newblock

\PrintBackRefs{\CurrentBib}

\bibitem [\protect \citeauthoryear {%
M.%
, Harikrishnan%
\BCBL {}\ \BBA {} Ambika%
}{%
M.%
\ \protect \BOthers {.}}{%
{\protect \APACyear {2022}}%
}]{%
M.2022Recurrence}
\APACinsertmetastar {%
M.2022Recurrence}%
\begin{APACrefauthors}%
M., K.%
, Harikrishnan, K.P.%
\BCBL {} Ambika, G.%
\end{APACrefauthors}%
\unskip\
\newblock
\APACrefYearMonthDay{2022}{}{}.
\newblock
{\BBOQ}\APACrefatitle {Recurrence measures and transitions in stock market dynamics} {Recurrence measures and transitions in stock market dynamics}.{\BBCQ}
\newblock
\APACjournalVolNumPages{Physica A: Statistical Mechanics and its Applications}{}{}{,}
\newblock
\begin{APACrefDOI} \doi{10.1016/j.physa.2022.128240} \end{APACrefDOI}
\newblock

\newblock

\PrintBackRefs{\CurrentBib}

\bibitem [\protect \citeauthoryear {%
Maeso%
\ \BBA {} Martellini%
}{%
Maeso%
\ \BBA {} Martellini%
}{%
{\protect \APACyear {2020}}%
}]{%
Maeso2020Maximizing}
\APACinsertmetastar {%
Maeso2020Maximizing}%
\begin{APACrefauthors}%
Maeso, J.%
\BCBT {}\ \BBA {} Martellini, L.%
\end{APACrefauthors}%
\unskip\
\newblock
\APACrefYearMonthDay{2020}{}{}.
\newblock
{\BBOQ}\APACrefatitle {Maximizing an equity portfolio excess growth rate: a new form of smart beta strategy?} {Maximizing an equity portfolio excess growth rate: a new form of smart beta strategy?}{\BBCQ}
\newblock
\APACjournalVolNumPages{Quantitative Finance}{20}{}{1185 - 1197,}
\newblock
\begin{APACrefDOI} \doi{10.1080/14697688.2020.1729398} \end{APACrefDOI}
\newblock

\newblock

\PrintBackRefs{\CurrentBib}

\bibitem [\protect \citeauthoryear {%
Mousavi%
\ \BBA {} Shen%
}{%
Mousavi%
\ \BBA {} Shen%
}{%
{\protect \APACyear {2021}}%
}]{%
Mousavi2021A}
\APACinsertmetastar {%
Mousavi2021A}%
\begin{APACrefauthors}%
Mousavi, A.%
\BCBT {}\ \BBA {} Shen, J.%
\end{APACrefauthors}%
\unskip\
\newblock
\APACrefYearMonthDay{2021}{}{}.
\newblock
{\BBOQ}\APACrefatitle {A penalty decomposition algorithm with greedy improvement for mean‐reverting portfolios with sparsity and volatility constraints} {A penalty decomposition algorithm with greedy improvement for mean‐reverting portfolios with sparsity and volatility constraints}.{\BBCQ}
\newblock
\APACjournalVolNumPages{International Transactions in Operational Research}{}{}{,}
\newblock
\begin{APACrefDOI} \doi{10.1111/itor.13123} \end{APACrefDOI}
\newblock

\newblock

\PrintBackRefs{\CurrentBib}

\bibitem [\protect \citeauthoryear {%
Murphy%
}{%
Murphy%
}{%
{\protect \APACyear {1999}}%
}]{%
Murphy1999}
\APACinsertmetastar {%
Murphy1999}%
\begin{APACrefauthors}%
Murphy, J.J.%
\end{APACrefauthors}%
\unskip\
\newblock
\APACrefYearMonthDay{1999}{}{}.
\newblock
\APACrefbtitle {Technical analysis of the financial markets} {Technical analysis of the financial markets}\ (\BVOL~77).
\PrintBackRefs{\CurrentBib}

\bibitem [\protect \citeauthoryear {%
Naeem%
, Rizvi%
\BCBL {}\ \BBA {} Coronato%
}{%
Naeem%
\ \protect \BOthers {.}}{%
{\protect \APACyear {2020}}%
}]{%
Naeem2020A}
\APACinsertmetastar {%
Naeem2020A}%
\begin{APACrefauthors}%
Naeem, M.%
, Rizvi, S.T.H.%
\BCBL {} Coronato, A.%
\end{APACrefauthors}%
\unskip\
\newblock
\APACrefYearMonthDay{2020}{}{}.
\newblock
{\BBOQ}\APACrefatitle {A Gentle Introduction to Reinforcement Learning and its Application in Different Fields} {A gentle introduction to reinforcement learning and its application in different fields}.{\BBCQ}
\newblock
\APACjournalVolNumPages{IEEE Access}{8}{}{209320-209344,}
\newblock
\begin{APACrefDOI} \doi{10.1109/ACCESS.2020.3038605} \end{APACrefDOI}
\newblock

\newblock

\PrintBackRefs{\CurrentBib}

\bibitem [\protect \citeauthoryear {%
Nasir%
\ \BBA {} Soliman%
}{%
Nasir%
\ \BBA {} Soliman%
}{%
{\protect \APACyear {2014}}%
}]{%
Nasir2014Aspects}
\APACinsertmetastar {%
Nasir2014Aspects}%
\begin{APACrefauthors}%
Nasir, M.%
\BCBT {}\ \BBA {} Soliman, A.M.%
\end{APACrefauthors}%
\unskip\
\newblock
\APACrefYearMonthDay{2014}{}{}.
\newblock
{\BBOQ}\APACrefatitle {Aspects of Macroeconomic Policy Combinations and Their Effects on Financial Markets} {Aspects of macroeconomic policy combinations and their effects on financial markets}.{\BBCQ}
\newblock
\APACjournalVolNumPages{Banking\& Insurance eJournal}{}{}{,}
\newblock

\newblock

\PrintBackRefs{\CurrentBib}

\bibitem [\protect \citeauthoryear {%
Onireti%
\ \protect \BOthers {.}}{%
Onireti%
\ \protect \BOthers {.}}{%
{\protect \APACyear {2016}}%
}]{%
Onireti2016}
\APACinsertmetastar {%
Onireti2016}%
\begin{APACrefauthors}%
Onireti, O.%
, Zoha, A.%
, Moysen, J.%
, Imran, A.%
, Giupponi, L.%
, Imran, M.A.%
\BCBL {} Abu-Dayya, A.%
\end{APACrefauthors}%
\unskip\
\newblock
\APACrefYearMonthDay{2016}{}{}.
\newblock
{\BBOQ}\APACrefatitle {A cell outage management framework for dense heterogeneous networks} {A cell outage management framework for dense heterogeneous networks}.{\BBCQ}
\newblock
\APACjournalVolNumPages{IEEE Transactions on Vehicular Technology}{65}{}{,}
\newblock
\begin{APACrefDOI} \doi{10.1109/TVT.2015.2431371} \end{APACrefDOI}
\newblock

\newblock

\PrintBackRefs{\CurrentBib}

\bibitem [\protect \citeauthoryear {%
Parisi%
}{%
Parisi%
}{%
{\protect \APACyear {2020}}%
}]{%
Parisi2020Reinforcement}
\APACinsertmetastar {%
Parisi2020Reinforcement}%
\begin{APACrefauthors}%
Parisi, S.%
\end{APACrefauthors}%
\unskip\
\newblock
\APACrefYearMonthDay{2020}{}{}.
\newblock
{\BBOQ}\APACrefatitle {Reinforcement Learning with Sparse and Multiple Rewards} {Reinforcement learning with sparse and multiple rewards}.{\BBCQ}
\newblock

\newblock
\begin{APACrefDOI} \doi{10.25534/TUPRINTS-00011372} \end{APACrefDOI}
\newblock

\newblock

\PrintBackRefs{\CurrentBib}

\bibitem [\protect \citeauthoryear {%
Paul%
\ \protect \BOthers {.}}{%
Paul%
\ \protect \BOthers {.}}{%
{\protect \APACyear {2018}}%
}]{%
2018CONTINUOUS}
\APACinsertmetastar {%
2018CONTINUOUS}%
\begin{APACrefauthors}%
Paul, L.T.%
, James, H.J.%
, David, S.%
, Tom, E.%
, Yuval, T.%
, Otto, H.N.M.%
\BDBL {}Alexander, P.%
\end{APACrefauthors}%
\unskip\
\newblock
\APACrefYearMonthDay{2018}{}{}.
\newblock
{\BBOQ}\APACrefatitle {CONTINUOUS CONTROL WITH DEEP REINFORCEMENT LEARNING} {Continuous control with deep reinforcement learning}.{\BBCQ}
\newblock

\newblock

\newblock

\PrintBackRefs{\CurrentBib}

\bibitem [\protect \citeauthoryear {%
Paule-Vianez%
, Gómez-Martínez%
\BCBL {}\ \BBA {} Prado‐Román%
}{%
Paule-Vianez%
\ \protect \BOthers {.}}{%
{\protect \APACyear {2020}}%
}]{%
Paule-Vianez2020A}
\APACinsertmetastar {%
Paule-Vianez2020A}%
\begin{APACrefauthors}%
Paule-Vianez, J.%
, Gómez-Martínez, R.%
\BCBL {} Prado‐Román, C.%
\end{APACrefauthors}%
\unskip\
\newblock
\APACrefYearMonthDay{2020}{}{}.
\newblock
{\BBOQ}\APACrefatitle {A bibliometric analysis of behavioural finance with mapping analysis tools} {A bibliometric analysis of behavioural finance with mapping analysis tools}.{\BBCQ}
\newblock
\APACjournalVolNumPages{European Research on Management and Business Economics}{26}{}{71-77,}
\newblock
\begin{APACrefDOI} \doi{10.1016/j.iedeen.2020.01.001} \end{APACrefDOI}
\newblock

\newblock

\PrintBackRefs{\CurrentBib}

\bibitem [\protect \citeauthoryear {%
Poon%
\ \BBA {} Granger%
}{%
Poon%
\ \BBA {} Granger%
}{%
{\protect \APACyear {2003}}%
}]{%
Poon2003}
\APACinsertmetastar {%
Poon2003}%
\begin{APACrefauthors}%
Poon, S.H.%
\BCBT {}\ \BBA {} Granger, C.W.%
\end{APACrefauthors}%
\unskip\
\newblock
\APACrefYearMonthDay{2003}{}{}.
\newblock
\APACrefbtitle {Forecasting volatility in financial markets: A review} {Forecasting volatility in financial markets: A review}\ (\BVOL~41).
\PrintBackRefs{\CurrentBib}

\bibitem [\protect \citeauthoryear {%
Qin%
, Gu%
\BCBL {}\ \BBA {} Su%
}{%
Qin%
\ \protect \BOthers {.}}{%
{\protect \APACyear {2022}}%
}]{%
Qin2022}
\APACinsertmetastar {%
Qin2022}%
\begin{APACrefauthors}%
Qin, Y.%
, Gu, F.%
\BCBL {} Su, J.%
\end{APACrefauthors}%
\unskip\
\newblock
\APACrefYearMonthDay{2022}{}{}.
\newblock
{\BBOQ}\APACrefatitle {A Novel Deep Reinforcement Learning Strategy for Portfolio Management} {A novel deep reinforcement learning strategy for portfolio management}.{\BBCQ}.
\PrintBackRefs{\CurrentBib}

\bibitem [\protect \citeauthoryear {%
Ren%
, Jiang%
\BCBL {}\ \BBA {} Su%
}{%
Ren%
\ \protect \BOthers {.}}{%
{\protect \APACyear {2021}}%
}]{%
Ren2021}
\APACinsertmetastar {%
Ren2021}%
\begin{APACrefauthors}%
Ren, X.%
, Jiang, Z.%
\BCBL {} Su, J.%
\end{APACrefauthors}%
\unskip\
\newblock
\APACrefYearMonthDay{2021}{}{}.
\newblock
{\BBOQ}\APACrefatitle {The Use of Features to Enhance the Capability of Deep Reinforcement Learning for Investment Portfolio Management} {The use of features to enhance the capability of deep reinforcement learning for investment portfolio management}.{\BBCQ}.
\PrintBackRefs{\CurrentBib}

\bibitem [\protect \citeauthoryear {%
Rollinger%
\ \BBA {} Hoffman%
}{%
Rollinger%
\ \BBA {} Hoffman%
}{%
{\protect \APACyear {2015}}%
}]{%
Rollinger2015}
\APACinsertmetastar {%
Rollinger2015}%
\begin{APACrefauthors}%
Rollinger, T.N.%
\BCBT {}\ \BBA {} Hoffman, S.T.%
\end{APACrefauthors}%
\unskip\
\newblock
\APACrefYearMonthDay{2015}{}{}.
\newblock
{\BBOQ}\APACrefatitle {Sortino A Sharper Ratio} {Sortino a sharper ratio}.{\BBCQ}
\newblock
\APACjournalVolNumPages{Red Rock Capital}{}{}{,}
\newblock

\newblock

\PrintBackRefs{\CurrentBib}

\bibitem [\protect \citeauthoryear {%
Schulman%
, Wolski%
, Dhariwal%
, Radford%
\BCBL {}\ \BBA {} Klimov%
}{%
Schulman%
\ \protect \BOthers {.}}{%
{\protect \APACyear {2017}}%
}]{%
Schulman2017}
\APACinsertmetastar {%
Schulman2017}%
\begin{APACrefauthors}%
Schulman, J.%
, Wolski, F.%
, Dhariwal, P.%
, Radford, A.%
\BCBL {} Klimov, O.%
\end{APACrefauthors}%
\unskip\
\newblock
\APACrefYearMonthDay{2017}{}{}.
\newblock
{\BBOQ}\APACrefatitle {Proximal Policy Optimization Algorithms} {Proximal policy optimization algorithms}.{\BBCQ}
\newblock

\newblock
\begin{APACrefDOI} \doi{10.48550} \end{APACrefDOI}
\newblock

\newblock

\PrintBackRefs{\CurrentBib}

\bibitem [\protect \citeauthoryear {%
Sharif%
, Aloui%
\BCBL {}\ \BBA {} Yarovaya%
}{%
Sharif%
\ \protect \BOthers {.}}{%
{\protect \APACyear {2020}}%
}]{%
Sharif2020COVID-19}
\APACinsertmetastar {%
Sharif2020COVID-19}%
\begin{APACrefauthors}%
Sharif, A.%
, Aloui, C.%
\BCBL {} Yarovaya, L.%
\end{APACrefauthors}%
\unskip\
\newblock
\APACrefYearMonthDay{2020}{}{}.
\newblock
{\BBOQ}\APACrefatitle {COVID-19 pandemic, oil prices, stock market, geopolitical risk and policy uncertainty nexus in the US economy: Fresh evidence from the wavelet-based approach} {Covid-19 pandemic, oil prices, stock market, geopolitical risk and policy uncertainty nexus in the us economy: Fresh evidence from the wavelet-based approach}.{\BBCQ}
\newblock
\APACjournalVolNumPages{International Review of Financial Analysis}{70}{}{101496 - 101496,}
\newblock
\begin{APACrefDOI} \doi{10.1016/j.irfa.2020.101496} \end{APACrefDOI}
\newblock

\newblock

\PrintBackRefs{\CurrentBib}

\bibitem [\protect \citeauthoryear {%
Sharpe%
}{%
Sharpe%
}{%
{\protect \APACyear {1994}}%
}]{%
Sharpe1994}
\APACinsertmetastar {%
Sharpe1994}%
\begin{APACrefauthors}%
Sharpe, W.F.%
\end{APACrefauthors}%
\unskip\
\newblock
\APACrefYearMonthDay{1994}{}{}.
\newblock
{\BBOQ}\APACrefatitle {The Sharpe Ratio} {The sharpe ratio}.{\BBCQ}
\newblock
\APACjournalVolNumPages{The Journal of Portfolio Management}{21}{}{,}
\newblock
\begin{APACrefDOI} \doi{10.3905/jpm.1994.409501} \end{APACrefDOI}
\newblock

\newblock

\PrintBackRefs{\CurrentBib}

\bibitem [\protect \citeauthoryear {%
Singleton%
}{%
Singleton%
}{%
{\protect \APACyear {2014}}%
}]{%
Singleton2014}
\APACinsertmetastar {%
Singleton2014}%
\begin{APACrefauthors}%
Singleton, K.J.%
\end{APACrefauthors}%
\unskip\
\newblock
\APACrefYearMonthDay{2014}{}{}.
\newblock
{\BBOQ}\APACrefatitle {Investor flows and the 2008 boom/bust in oil prices} {Investor flows and the 2008 boom/bust in oil prices}.{\BBCQ}
\newblock
\APACjournalVolNumPages{Management Science}{60}{}{,}
\newblock
\begin{APACrefDOI} \doi{10.1287/mnsc.2013.1756} \end{APACrefDOI}
\newblock

\newblock

\PrintBackRefs{\CurrentBib}

\bibitem [\protect \citeauthoryear {%
Soleymani%
\ \BBA {} Paquet%
}{%
Soleymani%
\ \BBA {} Paquet%
}{%
{\protect \APACyear {2020}}%
}]{%
Soleymani2020}
\APACinsertmetastar {%
Soleymani2020}%
\begin{APACrefauthors}%
Soleymani, F.%
\BCBT {}\ \BBA {} Paquet, E.%
\end{APACrefauthors}%
\unskip\
\newblock
\APACrefYearMonthDay{2020}{}{}.
\newblock
{\BBOQ}\APACrefatitle {Financial portfolio optimization with online deep reinforcement learning and restricted stacked autoencoder—DeepBreath} {Financial portfolio optimization with online deep reinforcement learning and restricted stacked autoencoder—deepbreath}.{\BBCQ}
\newblock
\APACjournalVolNumPages{Expert Systems with Applications}{156}{}{,}
\newblock
\begin{APACrefDOI} \doi{10.1016/j.eswa.2020.113456} \end{APACrefDOI}
\newblock

\newblock

\PrintBackRefs{\CurrentBib}

\bibitem [\protect \citeauthoryear {%
Song%
, Jiang%
, Tu%
, Du%
\BCBL {}\ \BBA {} Neyshabur%
}{%
Song%
\ \protect \BOthers {.}}{%
{\protect \APACyear {2019}}%
}]{%
Song2019Observational}
\APACinsertmetastar {%
Song2019Observational}%
\begin{APACrefauthors}%
Song, X.%
, Jiang, Y.%
, Tu, S.%
, Du, Y.%
\BCBL {} Neyshabur, B.%
\end{APACrefauthors}%
\unskip\
\newblock
\APACrefYearMonthDay{2019}{}{}.
\newblock
{\BBOQ}\APACrefatitle {Observational Overfitting in Reinforcement Learning} {Observational overfitting in reinforcement learning}.{\BBCQ}
\newblock
\APACjournalVolNumPages{ArXiv}{abs/1912.02975}{}{,}
\newblock

\newblock

\PrintBackRefs{\CurrentBib}

\bibitem [\protect \citeauthoryear {%
R.~Sun%
, Jiang%
\BCBL {}\ \BBA {} Su%
}{%
R.~Sun%
\ \protect \BOthers {.}}{%
{\protect \APACyear {2021}}%
}]{%
Sun2021}
\APACinsertmetastar {%
Sun2021}%
\begin{APACrefauthors}%
Sun, R.%
, Jiang, Z.%
\BCBL {} Su, J.%
\end{APACrefauthors}%
\unskip\
\newblock
\APACrefYearMonthDay{2021}{}{}.
\newblock
{\BBOQ}\APACrefatitle {A Deep Residual Shrinkage Neural Network-based Deep Reinforcement Learning Strategy in Financial Portfolio Management} {A deep residual shrinkage neural network-based deep reinforcement learning strategy in financial portfolio management}.{\BBCQ}.
\PrintBackRefs{\CurrentBib}

\bibitem [\protect \citeauthoryear {%
S.~Sun%
\ \protect \BOthers {.}}{%
S.~Sun%
\ \protect \BOthers {.}}{%
{\protect \APACyear {2023}}%
}]{%
sun2023trademaster}
\APACinsertmetastar {%
sun2023trademaster}%
\begin{APACrefauthors}%
Sun, S.%
, Qin, M.%
, Xia, H.%
, Zong, C.%
, Ying, J.%
, Xie, Y.%
\BDBL {}others%
\end{APACrefauthors}%
\unskip\
\newblock
\APACrefYearMonthDay{2023}{}{}.
\newblock
{\BBOQ}\APACrefatitle {TradeMaster: A Holistic Quantitative Trading Platform Empowered by Reinforcement Learning} {Trademaster: A holistic quantitative trading platform empowered by reinforcement learning}.{\BBCQ}
\newblock
 \APACrefbtitle {Thirty-seventh Conference on Neural Information Processing Systems Datasets and Benchmarks Track.} {Thirty-seventh conference on neural information processing systems datasets and benchmarks track.}
\PrintBackRefs{\CurrentBib}

\bibitem [\protect \citeauthoryear {%
Sutton%
\ \BBA {} Barto%
}{%
Sutton%
\ \BBA {} Barto%
}{%
{\protect \APACyear {1998}}%
}]{%
1998Reinforcement}
\APACinsertmetastar {%
1998Reinforcement}%
\begin{APACrefauthors}%
Sutton, R.S.%
\BCBT {}\ \BBA {} Barto, A.G.%
\end{APACrefauthors}%
\unskip\
\newblock
\APACrefYearMonthDay{1998}{}{}.
\newblock
{\BBOQ}\APACrefatitle {Reinforcement Learning} {Reinforcement learning}.{\BBCQ}
\newblock
\APACjournalVolNumPages{A Bradford Book}{volume 15}{7}{665-685,}
\newblock

\newblock

\PrintBackRefs{\CurrentBib}

\bibitem [\protect \citeauthoryear {%
Takada%
\ \BBA {} Kitajima%
}{%
Takada%
\ \BBA {} Kitajima%
}{%
{\protect \APACyear {2022}}%
}]{%
Takada2022Trend-following}
\APACinsertmetastar {%
Takada2022Trend-following}%
\begin{APACrefauthors}%
Takada, T.%
\BCBT {}\ \BBA {} Kitajima, T.%
\end{APACrefauthors}%
\unskip\
\newblock
\APACrefYearMonthDay{2022}{}{}.
\newblock
{\BBOQ}\APACrefatitle {Trend-following with better adaptation to large downside risks} {Trend-following with better adaptation to large downside risks}.{\BBCQ}
\newblock
\APACjournalVolNumPages{PLoS ONE}{17}{}{,}
\newblock
\begin{APACrefDOI} \doi{10.1371/journal.pone.0276322} \end{APACrefDOI}
\newblock

\newblock

\PrintBackRefs{\CurrentBib}

\bibitem [\protect \citeauthoryear {%
Ulyanov%
, Vedaldi%
\BCBL {}\ \BBA {} Lempitsky%
}{%
Ulyanov%
\ \protect \BOthers {.}}{%
{\protect \APACyear {2014}}%
}]{%
Ulyanov2014}
\APACinsertmetastar {%
Ulyanov2014}%
\begin{APACrefauthors}%
Ulyanov, D.%
, Vedaldi, A.%
\BCBL {} Lempitsky, V.%
\end{APACrefauthors}%
\unskip\
\newblock
\APACrefYearMonthDay{2014}{}{}.
\newblock
{\BBOQ}\APACrefatitle {Instance Normalization: The Missing Ingredient for Fast Stylization Dmitry} {Instance normalization: The missing ingredient for fast stylization dmitry}.{\BBCQ}
\newblock
\APACjournalVolNumPages{Physical Review D - Particles, Fields, Gravitation and Cosmology}{}{}{,}
\newblock

\newblock

\PrintBackRefs{\CurrentBib}

\bibitem [\protect \citeauthoryear {%
Umar%
, Adekoya%
, Oliyide%
\BCBL {}\ \BBA {} Gubareva%
}{%
Umar%
\ \protect \BOthers {.}}{%
{\protect \APACyear {2021}}%
}]{%
Umar2021Media}
\APACinsertmetastar {%
Umar2021Media}%
\begin{APACrefauthors}%
Umar, Z.%
, Adekoya, O.%
, Oliyide, J.%
\BCBL {} Gubareva, M.%
\end{APACrefauthors}%
\unskip\
\newblock
\APACrefYearMonthDay{2021}{}{}.
\newblock
{\BBOQ}\APACrefatitle {Media sentiment and short stocks performance during a systemic crisis} {Media sentiment and short stocks performance during a systemic crisis}.{\BBCQ}
\newblock
\APACjournalVolNumPages{International Review of Financial Analysis}{}{}{,}
\newblock
\begin{APACrefDOI} \doi{10.1016/j.irfa.2021.101896} \end{APACrefDOI}
\newblock

\newblock

\PrintBackRefs{\CurrentBib}

\bibitem [\protect \citeauthoryear {%
Vaswani%
\ \protect \BOthers {.}}{%
Vaswani%
\ \protect \BOthers {.}}{%
{\protect \APACyear {2017}}%
}]{%
Vaswani2017}
\APACinsertmetastar {%
Vaswani2017}%
\begin{APACrefauthors}%
Vaswani, A.%
, Shazeer, N.%
, Parmar, N.%
, Uszkoreit, J.%
, Jones, L.%
, Gomez, A.N.%
\BDBL {}Polosukhin, I.%
\end{APACrefauthors}%
\unskip\
\newblock
\APACrefYearMonthDay{2017}{}{}.
\newblock
{\BBOQ}\APACrefatitle {Attention is all you need} {Attention is all you need}.{\BBCQ}
\newblock
 (\BVOL\ 2017-December).
\PrintBackRefs{\CurrentBib}

\bibitem [\protect \citeauthoryear {%
Vázquez-Canteli%
\ \BBA {} Nagy%
}{%
Vázquez-Canteli%
\ \BBA {} Nagy%
}{%
{\protect \APACyear {2019}}%
}]{%
Jos2019}
\APACinsertmetastar {%
Jos2019}%
\begin{APACrefauthors}%
Vázquez-Canteli, J.R.%
\BCBT {}\ \BBA {} Nagy, Z.%
\end{APACrefauthors}%
\unskip\
\newblock
\APACrefYearMonthDay{2019}{}{}.
\newblock
\APACrefbtitle {Reinforcement learning for demand response: A review of algorithms and modeling techniques} {Reinforcement learning for demand response: A review of algorithms and modeling techniques}\ (\BVOL~235).
\PrintBackRefs{\CurrentBib}

\bibitem [\protect \citeauthoryear {%
Wang%
, Yan%
\BCBL {}\ \BBA {} Zheng%
}{%
Wang%
\ \protect \BOthers {.}}{%
{\protect \APACyear {2020}}%
}]{%
Wang2020Shorting}
\APACinsertmetastar {%
Wang2020Shorting}%
\begin{APACrefauthors}%
Wang, X.%
, Yan, X.S.%
\BCBL {} Zheng, L.%
\end{APACrefauthors}%
\unskip\
\newblock
\APACrefYearMonthDay{2020}{}{}.
\newblock
{\BBOQ}\APACrefatitle {Shorting flows, public disclosure, and market efficiency} {Shorting flows, public disclosure, and market efficiency}.{\BBCQ}
\newblock
\APACjournalVolNumPages{Journal of Financial Economics}{135}{}{191-212,}
\newblock
\begin{APACrefDOI} \doi{10.1016/J.JFINECO.2019.05.018} \end{APACrefDOI}
\newblock

\newblock

\PrintBackRefs{\CurrentBib}

\bibitem [\protect \citeauthoryear {%
Whiteson%
, Tanner%
, Taylor%
\BCBL {}\ \BBA {} Stone%
}{%
Whiteson%
\ \protect \BOthers {.}}{%
{\protect \APACyear {2011}}%
}]{%
Whiteson2011Protecting}
\APACinsertmetastar {%
Whiteson2011Protecting}%
\begin{APACrefauthors}%
Whiteson, S.%
, Tanner, B.%
, Taylor, M.E.%
\BCBL {} Stone, P.%
\end{APACrefauthors}%
\unskip\
\newblock
\APACrefYearMonthDay{2011}{}{}.
\newblock
{\BBOQ}\APACrefatitle {Protecting against evaluation overfitting in empirical reinforcement learning} {Protecting against evaluation overfitting in empirical reinforcement learning}.{\BBCQ}
\newblock
\APACjournalVolNumPages{2011 IEEE Symposium on Adaptive Dynamic Programming and Reinforcement Learning (ADPRL)}{}{}{120-127,}
\newblock
\begin{APACrefDOI} \doi{10.1109/ADPRL.2011.5967363} \end{APACrefDOI}
\newblock

\newblock

\PrintBackRefs{\CurrentBib}

\bibitem [\protect \citeauthoryear {%
Xia%
, Yin%
, Dai%
\BCBL {}\ \BBA {} Jha%
}{%
Xia%
\ \protect \BOthers {.}}{%
{\protect \APACyear {2020}}%
}]{%
Xia2020Fully}
\APACinsertmetastar {%
Xia2020Fully}%
\begin{APACrefauthors}%
Xia, W.%
, Yin, H.%
, Dai, X.%
\BCBL {} Jha, N.%
\end{APACrefauthors}%
\unskip\
\newblock
\APACrefYearMonthDay{2020}{}{}.
\newblock
{\BBOQ}\APACrefatitle {Fully Dynamic Inference With Deep Neural Networks} {Fully dynamic inference with deep neural networks}.{\BBCQ}
\newblock
\APACjournalVolNumPages{IEEE Transactions on Emerging Topics in Computing}{10}{}{962-972,}
\newblock
\begin{APACrefDOI} \doi{10.1109/TETC.2021.3056031} \end{APACrefDOI}
\newblock

\newblock

\PrintBackRefs{\CurrentBib}

\bibitem [\protect \citeauthoryear {%
H.~Yang%
, Liu%
, Zhong%
\BCBL {}\ \BBA {} Walid%
}{%
H.~Yang%
\ \protect \BOthers {.}}{%
{\protect \APACyear {2020}}%
}]{%
Yang2020}
\APACinsertmetastar {%
Yang2020}%
\begin{APACrefauthors}%
Yang, H.%
, Liu, X\BHBI Y.%
, Zhong, S.%
\BCBL {} Walid, A.%
\end{APACrefauthors}%
\unskip\
\newblock
\APACrefYearMonthDay{2020}{}{}.
\newblock
{\BBOQ}\APACrefatitle {Deep Reinforcement Learning for Automated Stock Trading: An Ensemble Strategy} {Deep reinforcement learning for automated stock trading: An ensemble strategy}.{\BBCQ}
\newblock
\APACjournalVolNumPages{SSRN Electronic Journal}{}{}{,}
\newblock
\begin{APACrefDOI} \doi{10.2139/ssrn.3690996} \end{APACrefDOI}
\newblock

\newblock

\PrintBackRefs{\CurrentBib}

\bibitem [\protect \citeauthoryear {%
X.~Yang%
\ \protect \BOthers {.}}{%
X.~Yang%
\ \protect \BOthers {.}}{%
{\protect \APACyear {2022}}%
}]{%
Yang2022}
\APACinsertmetastar {%
Yang2022}%
\begin{APACrefauthors}%
Yang, X.%
, Sun, R.%
, Ren, X.%
, Stefanidis, A.%
, Gu, F.%
\BCBL {} Su, J.%
\end{APACrefauthors}%
\unskip\
\newblock
\APACrefYearMonthDay{2022}{}{}.
\newblock
{\BBOQ}\APACrefatitle {Ghost Expectation Point with Deep Reinforcement Learning in Financial Portfolio Management} {Ghost expectation point with deep reinforcement learning in financial portfolio management}.{\BBCQ}
\newblock
 (\BPG~136-142).
\PrintBackRefs{\CurrentBib}

\bibitem [\protect \citeauthoryear {%
Yao%
, Ren%
\BCBL {}\ \BBA {} Su%
}{%
Yao%
\ \protect \BOthers {.}}{%
{\protect \APACyear {2022}}%
}]{%
Yao2022}
\APACinsertmetastar {%
Yao2022}%
\begin{APACrefauthors}%
Yao, W.%
, Ren, X.%
\BCBL {} Su, J.%
\end{APACrefauthors}%
\unskip\
\newblock
\APACrefYearMonthDay{2022}{}{}.
\newblock
{\BBOQ}\APACrefatitle {An Inception Network with Bottleneck Attention Module for Deep Reinforcement Learning Framework in Financial Portfolio Management} {An inception network with bottleneck attention module for deep reinforcement learning framework in financial portfolio management}.{\BBCQ}.
\PrintBackRefs{\CurrentBib}

\bibitem [\protect \citeauthoryear {%
Ye%
\ \protect \BOthers {.}}{%
Ye%
\ \protect \BOthers {.}}{%
{\protect \APACyear {2020}}%
}]{%
Ye2020ReinforcementLearningBP}
\APACinsertmetastar {%
Ye2020ReinforcementLearningBP}%
\begin{APACrefauthors}%
Ye, Y.%
, Pei, H.%
, Wang, B.%
, Chen, P\BHBI Y.%
, Zhu, Y.%
, Xiao, J.%
\BCBL {} Li, B.%
\end{APACrefauthors}%
\unskip\
\newblock
\APACrefYearMonthDay{2020}{}{}.
\newblock
{\BBOQ}\APACrefatitle {Reinforcement-Learning based Portfolio Management with Augmented Asset Movement Prediction States} {Reinforcement-learning based portfolio management with augmented asset movement prediction states}.{\BBCQ}
\newblock
\APACjournalVolNumPages{ArXiv}{abs/2002.05780}{}{,}
\newblock
\begin{APACrefURL} {https://api.semanticscholar.org/CorpusID:211126763} \end{APACrefURL}
\newblock

\newblock

\PrintBackRefs{\CurrentBib}

\bibitem [\protect \citeauthoryear {%
R.~Zhang%
\ \protect \BOthers {.}}{%
R.~Zhang%
\ \protect \BOthers {.}}{%
{\protect \APACyear {2022}}%
}]{%
Zhang2022}
\APACinsertmetastar {%
Zhang2022}%
\begin{APACrefauthors}%
Zhang, R.%
, Ren, X.%
, Gu, F.%
, Stefanidis, A.%
, Sun, R.%
\BCBL {} Su, J.%
\end{APACrefauthors}%
\unskip\
\newblock
\APACrefYearMonthDay{2022}{}{}.
\newblock
{\BBOQ}\APACrefatitle {MDAEN: Multi-Dimensional Attention-based Ensemble Network in Deep Reinforcement Learning Framework for Portfolio Management} {Mdaen: Multi-dimensional attention-based ensemble network in deep reinforcement learning framework for portfolio management}.{\BBCQ}
\newblock
 (\BPG~143-151).
\PrintBackRefs{\CurrentBib}

\bibitem [\protect \citeauthoryear {%
W.~Zhang%
, Chen%
, Yan%
, Zhang%
\BCBL {}\ \BBA {} Xu%
}{%
W.~Zhang%
\ \protect \BOthers {.}}{%
{\protect \APACyear {2021}}%
}]{%
Zhang2021A}
\APACinsertmetastar {%
Zhang2021A}%
\begin{APACrefauthors}%
Zhang, W.%
, Chen, Q.%
, Yan, J.%
, Zhang, S.%
\BCBL {} Xu, J.%
\end{APACrefauthors}%
\unskip\
\newblock
\APACrefYearMonthDay{2021}{}{}.
\newblock
{\BBOQ}\APACrefatitle {A novel asynchronous deep reinforcement learning model with adaptive early forecasting method and reward incentive mechanism for short-term load forecasting} {A novel asynchronous deep reinforcement learning model with adaptive early forecasting method and reward incentive mechanism for short-term load forecasting}.{\BBCQ}
\newblock
\APACjournalVolNumPages{Energy}{236}{}{121492,}
\newblock
\begin{APACrefDOI} \doi{10.1016/J.ENERGY.2021.121492} \end{APACrefDOI}
\newblock

\newblock

\PrintBackRefs{\CurrentBib}

\bibitem [\protect \citeauthoryear {%
Zhu%
\ \BBA {} Rosendo%
}{%
Zhu%
\ \BBA {} Rosendo%
}{%
{\protect \APACyear {2021}}%
}]{%
Zhu2021A}
\APACinsertmetastar {%
Zhu2021A}%
\begin{APACrefauthors}%
Zhu, W.%
\BCBT {}\ \BBA {} Rosendo, A.%
\end{APACrefauthors}%
\unskip\
\newblock
\APACrefYearMonthDay{2021}{}{}.
\newblock
{\BBOQ}\APACrefatitle {A Functional Clipping Approach for Policy Optimization Algorithms} {A functional clipping approach for policy optimization algorithms}.{\BBCQ}
\newblock
\APACjournalVolNumPages{IEEE Access}{9}{}{96056-96063,}
\newblock
\begin{APACrefDOI} \doi{10.1109/ACCESS.2021.3094566} \end{APACrefDOI}
\newblock

\newblock

\PrintBackRefs{\CurrentBib}

\end{thebibliography}

\end{document}